\documentclass[10pt,twocolumn,letterpaper]{article}

\usepackage{cvpr}              

\usepackage{graphicx}
\usepackage{amsmath}
\usepackage{amssymb}
\usepackage{booktabs}
\usepackage{times}
\usepackage{epsfig}
\usepackage{amsmath}
\usepackage{amssymb}
\usepackage{subfloat}
\usepackage{pifont}
\usepackage{makecell}
\usepackage[toc,page]{appendix}
\usepackage{color, colortbl}

\usepackage{makecell}
\usepackage[rightcaption]{sidecap} 
\usepackage{bm}
\usepackage{physics}
\usepackage{pbox}
\usepackage[mathscr]{eucal}
\usepackage{physics}
\usepackage{xcolor}
\usepackage{makecell}
\usepackage{algorithm} 
\usepackage{listings}

\def\R{\mathbb{R}}

\definecolor{lightsalmon}{rgb}{1.0, 0.63, 0.48}
\definecolor{navajowhite}{rgb}{1.0, 0.87, 0.68}
\definecolor{mediumspringbud}{rgb}{0.79, 0.86, 0.54}
\definecolor{gray}{rgb}{0.75, 0.75, 0.75}

\definecolor{moonstoneblue}{rgb}{0.45, 0.66, 0.76}

\usepackage[pagebackref,breaklinks,colorlinks]{hyperref}
\usepackage{graphics,color}
\usepackage{xcolor}
\definecolor{asparagus}{rgb}{0.53, 0.66, 0.42}
\definecolor{glaucous}{rgb}{0.38, 0.51, 0.71}
\definecolor{lighttaupe}{rgb}{0.7, 0.55, 0.43}
\definecolor{moonstoneblue}{rgb}{0.45, 0.66, 0.76}
\definecolor{satinsheengold}{rgb}{0.8, 0.63, 0.21}
\definecolor{tumbleweed}{rgb}{0.87, 0.67, 0.53}
\definecolor{wisteria}{rgb}{0.79, 0.63, 0.86}
\definecolor{wildblueyonder}{rgb}{0.64, 0.68, 0.82}
\definecolor{thulianpink}{rgb}{0.87, 0.44, 0.63}

\definecolor{olivine}{rgb}{0.6, 0.73, 0.45}
\definecolor{sapgreen}{rgb}{0.31, 0.49, 0.16}

\definecolor{arylideyellow}{rgb}{0.91, 0.84, 0.42}

\definecolor{bluebell}{rgb}{0.64, 0.64, 0.82}

\definecolor{junglegreen}{rgb}{0.16, 0.67, 0.53}

\definecolor{jonquil}{rgb}{0.98, 0.85, 0.37}

\definecolor{lightsalmonpink}{rgb}{1.0, 0.6, 0.6}

\definecolor{macaroniandcheese}{rgb}{1.0, 0.74, 0.53}

\definecolor{khaki(html/css)(khaki)}{rgb}{0.76, 0.69, 0.57}

\newsavebox{\ET}
\savebox{\ET}{\textcolor{asparagus}{\rule{1.5in}{1.5in}}}

\newsavebox{\WT}
\savebox{\WT}{\textcolor{glaucous}{\rule{1.5in}{1.5in}}}

\newsavebox{\TC}
\savebox{\TC}{\textcolor{lighttaupe}{\rule{1.5in}{1.5in}}}

\newsavebox{\Spleen}
\savebox{\Spleen}{\textcolor{wisteria}{\rule{1.5in}{1.5in}}}

\newsavebox{\RKID}
\savebox{\RKID}{\textcolor{khaki(html/css)(khaki)}{\rule{1.5in}{1.5in}}}

\newsavebox{\LKID}
\savebox{\LKID}{\textcolor{bluebell}{\rule{1.5in}{1.5in}}}

\newsavebox{\GALL}
\savebox{\GALL}{\textcolor{junglegreen}{\rule{1.5in}{1.5in}}}

\newsavebox{\LIVER}
\savebox{\LIVER}{\textcolor{satinsheengold}{\rule{1.5in}{1.5in}}}

\newsavebox{\STOMACH}
\savebox{\STOMACH}{\textcolor{lightsalmonpink}{\rule{1.5in}{1.5in}}}

\newsavebox{\AORTA}
\savebox{\AORTA}{\textcolor{macaroniandcheese}{\rule{1.5in}{1.5in}}}

\newsavebox{\PANC}
\savebox{\PANC}{\textcolor{olivine}{\rule{1.5in}{1.5in}}}

\newsavebox{\MSD}
\savebox{\MSD}{\textcolor{moonstoneblue}{\rule{1.5in}{1.5in}}}

\definecolor{label1}{HTML}{AFD4D5}

\definecolor{label2}{HTML}{C5D4AD}

\definecolor{label3}{HTML}{E3A6A3}

\definecolor{label4}{HTML}{FFFF00}

\definecolor{label5}{HTML}{84ACD8}

\definecolor{label6}{HTML}{DB9999}

\definecolor{label7}{HTML}{C199C3}

\definecolor{label8}{HTML}{6CB0CC}

\definecolor{label9}{HTML}{FCEF9F}

\definecolor{label10}{HTML}{BDF5BC}

\definecolor{label11}{HTML}{66CDAA}

\definecolor{label12}{HTML}{000080}

\definecolor{label13}{HTML}{EDFE85}

\newsavebox{\SPLEENN}
\savebox{\SPLEENN}{\textcolor{label1}{\rule{1.5in}{1.5in}}}

\newsavebox{\RIGHTKIDNEYY}
\savebox{\RIGHTKIDNEYY}{\textcolor{label2}{\rule{1.5in}{1.5in}}}

\newsavebox{\LEFTKIDNEYY}
\savebox{\LEFTKIDNEYY}{\textcolor{label3}{\rule{1.5in}{1.5in}}}

\newsavebox{\GALLBLADERR}
\savebox{\GALLBLADERR}{\textcolor{label4}{\rule{1.5in}{1.5in}}}

\newsavebox{\ESOPO}
\savebox{\ESOPO}{\textcolor{label5}{\rule{1.5in}{1.5in}}}

\newsavebox{\LIVERR}
\savebox{\LIVERR}{\textcolor{label6}{\rule{1.5in}{1.5in}}}

\newsavebox{\STOMACHH}
\savebox{\STOMACHH}{\textcolor{label7}{\rule{1.5in}{1.5in}}}

\newsavebox{\AORTAA}
\savebox{\AORTAA}{\textcolor{label8}{\rule{1.5in}{1.5in}}}

\newsavebox{\IVC}
\savebox{\IVC}{\textcolor{label9}{\rule{1.5in}{1.5in}}}

\newsavebox{\VEINS}
\savebox{\VEINS}{\textcolor{label10}{\rule{1.5in}{1.5in}}}

\newsavebox{\PANCREASS}
\savebox{\PANCREASS}{\textcolor{label11}{\rule{1.5in}{1.5in}}}

\newsavebox{\RADD}
\savebox{\RADD}{\textcolor{label12}{\rule{1.5in}{1.5in}}}

\newsavebox{\LADD}
\savebox{\LADD}{\textcolor{label13}{\rule{1.5in}{1.5in}}}

\usepackage[capitalize]{cleveref}
\crefname{section}{Sec.}{Secs.}
\Crefname{section}{Section}{Sections}
\Crefname{table}{Table}{Tables}
\crefname{table}{Tab.}{Tabs.}

\def\cvprPaperID{4415} 
\def\confName{WACV}
\def\confYear{2025}

\begin{document}

\title{Y-CA-Net: A Convolutional Attention Based Network for Volumetric Medical Image Segmentation}

\author{Muhammad Hamza Sharif, 
~~ Muzammal Naseer, 
~~ Mohammad Yaqub,
~~ Min Xu,
~~ Mohsen Guizani\\
Mohamed Bin Zayed University of Artificial Intelligence (MBZUAI)
}

\thispagestyle{empty}
\twocolumn[{
    \renewcommand\twocolumn[1][]{#1}
    \maketitle
{\includegraphics[trim={0.3cm 0.1cm 0 0},clip, width=\textwidth]{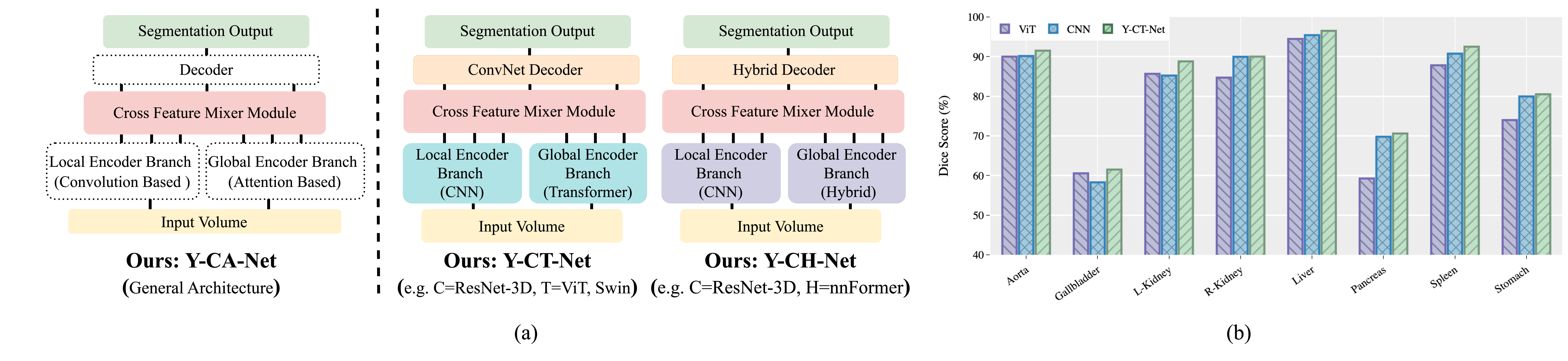}}\vspace{-1em}
\captionof{figure}{
    \footnotesize \textbf{Y-CA-Net and comparison of performance of Y-CA-Net based model with standalone convolutional and attention based model on Synapse dataset.} As shown in (a), we introduce \textit{Y-CA-Net} as a general architecture based upon two encoder branches, \ie convolutional and attention, a \emph{Cross Feature Mixer Module (CFMM)}, and a decoder. Our network, Y-CA-Net, is a hybrid architecture that leverages the strengths of both convolution and attention mechanisms and combines them through the CFMM before passing them to the decoder. We have presented two variants of our Y-shaped network, namely \textit{Y-CT-Net} and \textit{Y-CH-Net}, with different encoder and decoder backbones. We argue that for voxel-wise prediction tasks, preserving the local structure between neighborhood voxels is important, which is missed by the attention mechanism. To address this issue, we employ two encoders, a local encoder that relies solely on convolution-based operations and a global encoder that employs attention-based mechanisms. The features extracted from both encoders are then fused in the \emph{CFMM}. Remarkably, the resulting model, \textit{Y-CT-Net}, outperforms the well-tuned stand-alone vision Transformer baseline (UNETR) and CNN as shown in (b) which supports that Y-CA-Net provides competitive performance for volumetric segmentation.}\label{fig:concept-diagram}
\vspace{0.5em}}]

\begin{abstract}\vspace{-1.25em}
Recent attention-based volumetric segmentation (VS) methods have achieved remarkable performance in the medical domain which focuses on modeling long-range dependencies. However, for voxel-wise prediction tasks, discriminative local features are key components for the performance of the VS models which is missing in attention-based VS methods. Aiming at resolving this issue, we deliberately incorporate the convolutional encoder branch with transformer backbone to extract local and global features in a parallel manner and aggregate them in Cross Feature Mixer Module (CFMM) for better prediction of segmentation mask. Consequently, we observe that the derived model, Y-CT-Net, achieves competitive performance on multiple medical segmentation tasks. For example, on multi-organ segmentation, Y-CT-Net achieves an 82.4\% dice score, surpassing well-tuned VS Transformer/CNN-like baselines UNETR/ResNet-3D by 2.9\%/1.4\%. With the success of Y-CT-Net, we extend this concept with hybrid attention models, that derived Y-CH-Net model, which brings a 3\% improvement in terms of HD95 score for same segmentation task. The effectiveness of both models Y-CT-Net and Y-CH-Net verifies our hypothesis and motivates us to initiate the concept of ``Y-CA-Net", a versatile generic architecture based upon any two encoders and a decoder backbones, to fully exploit the complementary strengths of both convolution and attention mechanisms. Based on experimental results, we argue Y-CA-Net is a key player in achieving superior results for volumetric segmentation.
\vspace{-2em}
\end{abstract}

\section{Introduction}\label{sec:introduction}\vspace{-0.5em}

Significant progress has been made recently with the evolution of vision transformers (ViTs) for downstream tasks \cite{naseer2021intriguing, ranasinghe2021selfsupervised, gani2022train, khan2022transformers, zamir2022restormer}. With their consistently superior performance over convolutional neural networks (CNNs) in natural images, these architectures have also been adapted by the medical community \cite{saeed2022tmss, cao2021swin, dai2021transmed, Sharif_2022_BMVC} due to their exceptional characteristics of capturing pair-wise relations via the self-attention mechanism. Specifically, in downstream tasks concerning medical data, transformer-based backbones have extensively been studied for segmentation \cite{hatamizadeh2022unetr, cao2021swin, hatamizadeh2022swin, zhou2021nnformer, huang2021missformer, mtunet}. Recent approaches \cite{hatamizadeh2022swin, hatamizadeh2022unetr, zhou2021nnformer} designed for volumetric segmentation with these backbones (\eg ViT \cite{dosovitskiy2020image}, Swin\cite{swin2021} and nnFormer \cite{zhou2021nnformer}), rely on capturing features either by computing local, global or window based self-attention mechanism between the non-overlapping volumetric patches. As opposed to CNNs, these methods can learn long-range dependencies in earlier layers and show promising performance on medical image segmentation benchmark datasets. However, when it comes to challenging 3D segmentation which is specifically volumetric voxel-wise prediction, local information plays an important role, where the model focuses more on learning \textit{local} image structure \cite{bardes2022vicregl}. The core component used in these methods \cite{hatamizadeh2022unetr, hatamizadeh2022swin, zhou2021nnformer} for extracting features is solely based on attention (local, global, or window-based) mechanism which allows dynamic aggregation of features \cite{han2021demystifying}. For volumetric dense prediction, preserving the semantically important information in the immediate neighborhood of each voxel is very essential which is missed by the attention-based methods. Convolution is an appropriate operation for extracting local features and preserving spatial information which is not fully exploited in the above methods.

\begin{figure}[t]

\begin{minipage}{0.48\textwidth}
\centering
  \begin{minipage}{\linewidth}
\scalebox{0.05}{{\usebox{\SPLEENN}}} \scriptsize Spleen ~~~ \scalebox{0.05}{{\usebox{\RIGHTKIDNEYY}}} \scriptsize R-Kidney ~~~ \scalebox{0.05}{{\usebox{\LEFTKIDNEYY}}} \scriptsize L-Kidney ~~~ \scalebox{0.05}{{\usebox{\GALLBLADERR}}} \scriptsize Gallbladder ~~~ \scalebox{0.05}{{\usebox{\ESOPO}}} \scriptsize Esophagus~~~ \scalebox{0.05}{{\usebox{\LIVERR}}} \scriptsize Liver ~~~ \scalebox{0.05}{{\usebox{\STOMACHH}}} \scriptsize Stomach ~~~ \scalebox{0.05}{{\usebox{\AORTAA}}} \scriptsize Aorta ~~~ \scalebox{0.05}{{\usebox{\IVC}}} \scriptsize IVC ~~~ \scalebox{0.05}{{\usebox{\VEINS}}} \scriptsize Veins ~~~ \scalebox{0.05}{{\usebox{\PANCREASS}}} \scriptsize Pancreas ~~~ \scalebox{0.05}{{\usebox{\RADD}}} \scriptsize Rad ~~~ \scalebox{0.05}{{\usebox{\LADD}}} \scriptsize Lad
  \end{minipage}
\\
\vspace{0.2em}
  \begin{minipage}{0.24\textwidth}
      \centering
   \includegraphics[height=2.0cm, width=\linewidth]{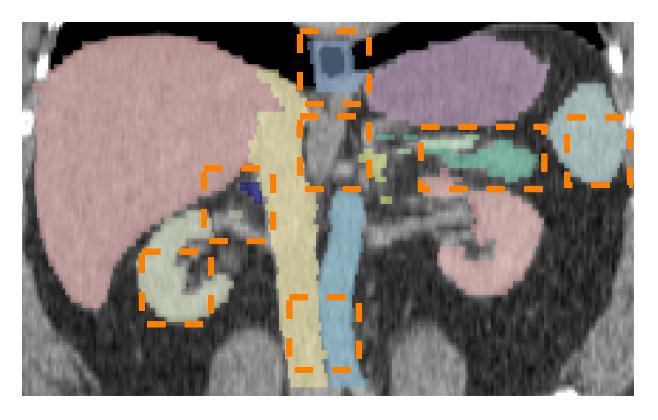}
  \end{minipage}
\begin{minipage}{0.24\textwidth}
      \centering
  \includegraphics[height=2.0cm, width=\linewidth]{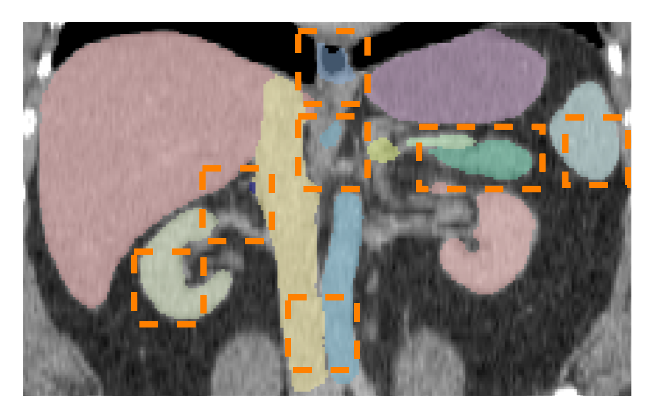}
  \end{minipage}
  \begin{minipage}{0.24\textwidth}
      \centering
    \includegraphics[height=2.0cm, width=\linewidth]{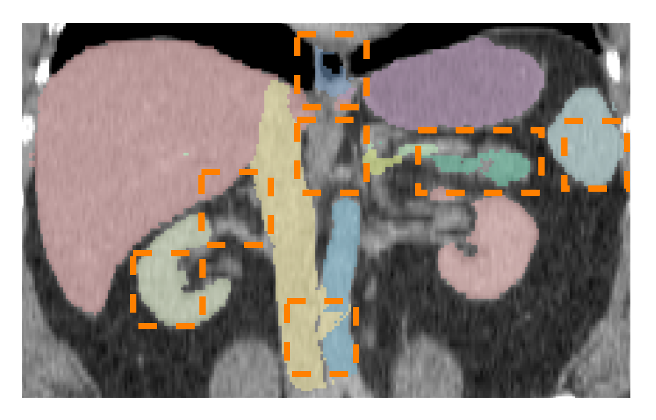}
  \end{minipage}
  \begin{minipage}{0.24\textwidth}
      \centering
   \includegraphics[height=2.0cm, width=\linewidth]{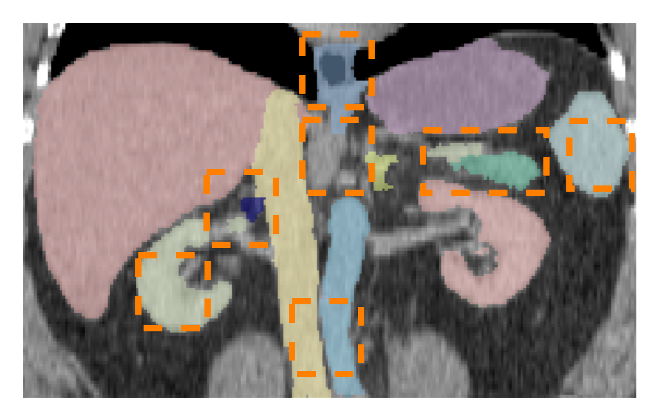}
  \end{minipage}
\\
\hspace{-0.3em}
 \begin{minipage}{0.24\textwidth}
      \centering
   \includegraphics[width=\linewidth]{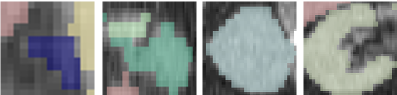}
   \footnotesize \textbf{Label}
  \end{minipage}
\begin{minipage}{0.24\textwidth}
      \centering
  \includegraphics[width=\linewidth]{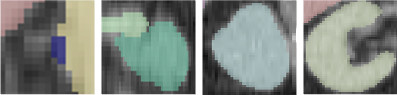}
 \footnotesize \textbf{CNN}
  \end{minipage}
  \begin{minipage}{0.24\textwidth}
      \centering
    \includegraphics[ width=\linewidth]{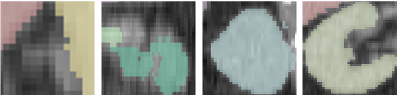}
    \footnotesize \textbf{ViT}
  \end{minipage}
  \begin{minipage}{0.24\textwidth}
      \centering
   \includegraphics[width=\linewidth]{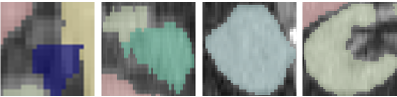}
   \footnotesize \textbf{Y-CT-Net}
  \end{minipage}

\vspace{-0.5em}
\caption{\footnotesize Visual comparison of transformer and ConvNet baseline with our Y-CT-Net on a multi-organ segmentation task. We observe some distinct differences that are highlighted by small square boxes. 
}\label{fig:comparsion}
\end{minipage}
\vspace{-1.5em}
\end{figure}

To verify our hypothesis that segmentation relies on discriminative local information, we compare the well-designed transformer-based method UNETR (using ViT backbone) \cite{hatamizadeh2022unetr} with standard ConvNet on the multi-organ task (see Sec. \ref{sec:methodology} and Fig. \ref{fig:concept-diagram} (a)). 
We observe the superiority of ConvNet over the transformer in most of the smaller organs but also highlight its inadequacy in cases where the transformer-based method has performed very well to segment the large organs as shown in Fig. \ref{fig:comparsion}. In light of the aforementioned issues, it makes us think that a combined approach is required to leverage the advantages of the transformers and CNNs and intelligently combine their learned representations. Therefore, we propose $\mathrm{\textbf{Y-CT-Net}}$, a Y-shaped architecture to combine the complementary strengths of both self-attention and convolution (see
Sec. \ref{sec:y-ct-net}). 
This derived model, termed $\mathrm{\textbf{Y-CT-Net}}$, achieves competitive performance, and even consistently outperforms well-tuned Transformer and standard ConvNet models as shown in Fig. \ref{fig:concept-diagram} (b) and Fig. \ref{fig:comparsion}. Specifically, Y-CT-Net performance on multi-organ tasks improves from 79.5\% to 82.4\% (\textit{dice score}) and 27.3\% to 19.5\% (\textit{hd95 score}). With the success of Y-CT-Net, we further extend this concept with hybrid attention models, that derive \textbf{Y-CH-Net} model (see Fig. \ref{fig:concept-diagram}(a)), resulting in a 3\% improvement in terms of \textit{hd95 score} for the same segmentation task (Table. \ref{table:multi-organ}). The efficacy of both models Y-CT-Net and Y-CH-Net supports our hypothesis and motivate us to develop the ``\textbf{Y-CA-Net}", a universal Y-shaped architecture that leverages the strengths of both \textbf{C}onvolution and \textbf{A}ttention mechanisms with any encoder decoder backbones. In summary, our contribution is three-folds:\vspace{-0.5em}
\begin{itemize}
\item We propose \textbf{Y-CA-Net} framework, a generic architecture with two encoder branches. Specifically, one of the encoder branch is purely convolution-based for learning local features, while the other encoder is attention-based for learning global representations. Our method aims to fully exploit the strengths of both convolution and attention mechanisms for volumetric segmentation (Sec. \ref{sec:y-ca-net}).\vspace{-0.5em}

\item We introduce \emph{Cross Feature Mixer Module (CFMM)} for learning semantically enriched features by mixing the local and global feature representations (Sec. \ref{sec:y-ca-net}). \vspace{-1.25em}

\item We introduce two variants of Y-CA-Net for volumetric segmentation which demonstrate significant performance improvements across various benchmark datasets for multi-organ and brain tumor segmentation as compared to existing state-of-the-art methods. \vspace{-0.5em}
\end{itemize}\vspace{-0.5em}

\section{Related Work}\vspace{-0.5em}
3D segmentation is a challenging task, especially in the medical imaging domain. Many studies have been proposed in the past that focus on either learning low-level or high-level features to predict segmentation masks. The inability of these architectures \cite{hatamizadeh2022unetr, unet, attunet, chen2021transunet, tang2022self, hatamizadeh2022swin} to capture both levels of features makes them an unfavorable choice for complex settings. In this regard, we discuss a few works on 3D biomedical image segmentation.\vspace{0.25em}

\noindent \textbf{Medical Segmentation with CNN-based Architectures:} In the domain of medical imaging, encoder-decoder style architectures are popular for segmentation tasks \cite{shamshad2022transformers}. With the advent of the U-Net model \cite{unet}, several CNN-based studies have been proposed such as U-Net++ \cite{unet++}, U-Net3++ \cite{Unet3+}, and 3D FCN \cite{FCN} that improve U-Net model \cite{unet} performance by utilizing dense connections between encoder and decoder. Subsequently, Zhou \etal \cite{zhou2018learning} proposed an ensemble approach by using multiple CNN-based backbones to incorporate multi-scale contextual information through an attention mechanism. Moreover, Isensee \etal \cite{nnunet} developed the nnU-Net model to demonstrate that vanilla U-Net architecture with minor modifications in a pre-processing stage is capable of achieving competitive performance in many medical imaging tasks.\vspace{0.25em}

\noindent \textbf{Medical Segmentation with Transformer Variants:}
The concept of the transformer was first introduced for sequence modeling tasks in the domain of natural language pre-processing \cite{vaswani2017attention}. Nowadays, transformers have made marvelous achievements in the field of computer vision and medical imaging for downstream tasks \cite{dosovitskiy2020image, liu2021swin, hatamizadeh2022unetr, tang2022self, hatamizadeh2022swin, zhou2021nnformer, saeed2022tmss}. Hatamizadeh \etal \cite{hatamizadeh2022unetr} proposed UNETR architecture, a 3D transformer with the ViT-based backbone for extracting the global features for volumetric medical segmentation tasks. With the achievement of the Swin transformer, Swin UNETR \cite{hatamizadeh2022swin} paper adopts the same idea of UNETR and uses Swin transformer backbone in the encoder branch for brain tumor segmentation tasks. Tang \etal \cite{tang2022self} extend Swin UNETR by incorporating self-supervised training strategy for multiple segmentation tasks. In addition to these, Swin-Unet \cite{cao2021swin} has also used swin transformer module to build U-shape architecture via skip connections for learning local and global representation for multi-organ segmentation tasks. However, despite all the significant efforts and achievements made by transformer-based methods for volumetric segmentation, these frameworks are still missing the element of local information which is significantly important for the voxel-wise prediction task.\vspace{0.25em}

\noindent \textbf{Hybrid Segmentation Methods:} Several recent works have explored hybrid architectures \cite{zhang2021transfuse, chen2021transunet, zhou2021nnformer, valanarasu2021medical} to combine convolution and self-attention operations in their backbones for better segmentation. TransFuse \cite{zhang2021transfuse} proposes a parallel CNN-transformer architecture with a BiFusion module to fuse multi-level features in encoder. Meanwhile, MedT \cite{valanarasu2021medical} introduces a gated position-sensitive axial-attention mechanism in self-attention to control the positional embedding information in the encoder and incorporates local-global strategy for training. Zhou \etal introduce an approach named nnFormer \cite{zhou2021nnformer} in which convolution layers transform the input scans into 3D patches, and self-attention modules build hierarchical feature pyramids. Although these methods have achieved promising performance, but they have not fully exploited the complementary strengths of both convolution and self-attention, which is necessary for the accurate segmentation of all types of organs.\vspace{-0.75em}

\section{Methodology}\label{sec:methodology}\vspace{-0.5em}
The ability to capture long-range dependencies or in other words modeling the interaction among input patches \cite{khan2022transformers, naseer2021intriguing} has been studied for the volumetric medical segmentation. The UNETR \cite{hatamizadeh2022unetr} and Swin UNETR \cite{hatamizadeh2022swin} were introduced as 3D transformer for the volumetric medical image segmentation. Their architecture follows an encoder-decoder network, with transformer backbones used in the encoder branch and convolutional layers in the decoder branch. Self-attention \cite{dosovitskiy2020image} is the workhorse of Vision Transformers-based encoders that capture global features and transfer these features to the decoder for segmentation. However, when it comes to voxel-wise prediction problems, local context information plays a significant role in the performance of the model as it focuses on the fine details within an image \cite{bardes2022vicregl}.  Modeling global interactions \cite{hatamizadeh2022unetr, hatamizadeh2022swin} does not effectively capture local details, which provide semantically meaningful information about voxel neighborhoods.\vspace{0.25em}

\noindent \textbf{Convolutional Vs. Transformer Encoders:}
In comparison to the traditional methods \cite{attunet, resunet}, the current approaches use a transformer-based encoder with a convolutional decoder for volumetric medical segmentation. Self-attention-based Transformers have complementary strengths as compared to convolutional filters. For example, Transformer architecture encodes longer-range dependencies while convolutional filter excels in extracting local features. Therefore, we first investigate if replacing the transformer with a purely convolutional encoder is less optimal. We first answer this question by directly training UNETR (after replacing its ViT-based encoder backbone with CNN) on a multi-organ dataset. We have directly utilized the first three stages of the convolutional encoder in our training pipeline for this experiment. More details of the convolutional encoder have been provided in the Sec. \ref{sec:localencoder}.\vspace{0.25em}

\noindent \textbf{Insights and Analysis:}
The results of UNETR with a CNN-based encoder are quite surprising and we found that it outperforms UNETR with ViT-based encoder. We observe that simply replacing the ViT-based encoder in UNETR \cite{hatamizadeh2022unetr} with a 3D CNN improves the segmentation performance which can be inferred from qualitative and quantitative results shown in Fig. \ref{fig:concept-diagram} (b) and \ref{fig:comparsion}. Although local features play a crucial role in voxel-level prediction, but visual results suggest that CNNs primarily focus on preserving local details in smaller organs (\eg pancreas, rad), while ViTs excel in retaining global information in larger organs (\eg right kidney, spleen) that may be neglected by CNNs. Inspired by the distinctive advantages of CNNs and ViTs in capturing local and global features, we aim to combine these features in such a way that model provides accurate prediction on small as well as large organs (Fig. \ref{fig:comparsion}).\vspace{-0.5em}

\setlength{\textfloatsep}{3pt}
\begin{algorithm}[t]
\caption{\footnotesize FeatureMixing for Y-CA-Net, PyTorch-like Code}
\label{alg:code}
\definecolor{codeblue}{rgb}{0.19, 0.55, 0.91}
\definecolor{codekw}{rgb}{0.77, 0.12, 0.23}
\lstset{
  backgroundcolor=\color{white},
  basicstyle=\fontsize{6.5pt}{6.5pt}\ttfamily\selectfont,
  keywordstyle=\bfseries,
  columns=fixed,
  keepspaces=true,
  breaklines=true,
  captionpos=b,
  commentstyle=\fontsize{6.5pt}{7.5pt}\color{codeblue},
  keywordstyle=\fontsize{6.5pt}{7.5pt}\color{codekw},
}
\begin{lstlisting}[language=python]
# Local Feature: fl , Global Feature: fg
import torch
import torch.nn as nn

class FeatureMixing(nn.Module):
    def __init__(self, in_channels):
        super(FeatureMixing, self).__init__()
        self.in_channels = in_channels
        
    def forward(self, fl, fg):
        # Check if spatial dimensions match
        if fl.size(-3) == fg.size(-3) and fl.size(-2) == 
        fg.size(-2) and fl.size(-1) == fg.size(-1):
            # Check if channel dimensions match
            if fl.size(1) != fg.size(1):
                # Apply projection layer to fl to change its channel dimension
                proj_layer = nn.Conv3d(fl.size(1), fg.size(1), kernel_size=1)
                fl = proj_layer(fl)
                
            # Add local and global features
            F_mix = torch.add(fl, fg)
            
        return F_mix
\end{lstlisting}
\end{algorithm}

\subsection{Y-CA-Net}\label{sec:y-ca-net}\vspace{-0.5em}
We first present the core concept of \textbf{Y-CA-Net} using Synapse dataset. 
As shown in Fig. \ref{fig:concept-diagram} (a), Y-CA-Net, is a general architecture based upon two encoder branches, \emph{Cross Feature Mixer Module (CFMM)}, and a decoder. The local encoder branch $\textbf{E}^L(.)$ is purely convolution based for extracting local features, while global encoder branch $\textbf{E}^G(.)$ is attention based for capturing global contextual information. The feature maps learned by these branches are passed through \emph{Cross Feature Mixer Module (CFMM)}, which intelligently aggregate them on the basis of their spatial dimension. These semantically enriched features are utilized by decoder for dense prediction of the segmentation mask.

\begin{figure*}
\begin{center}
\includegraphics[width=1\textwidth]{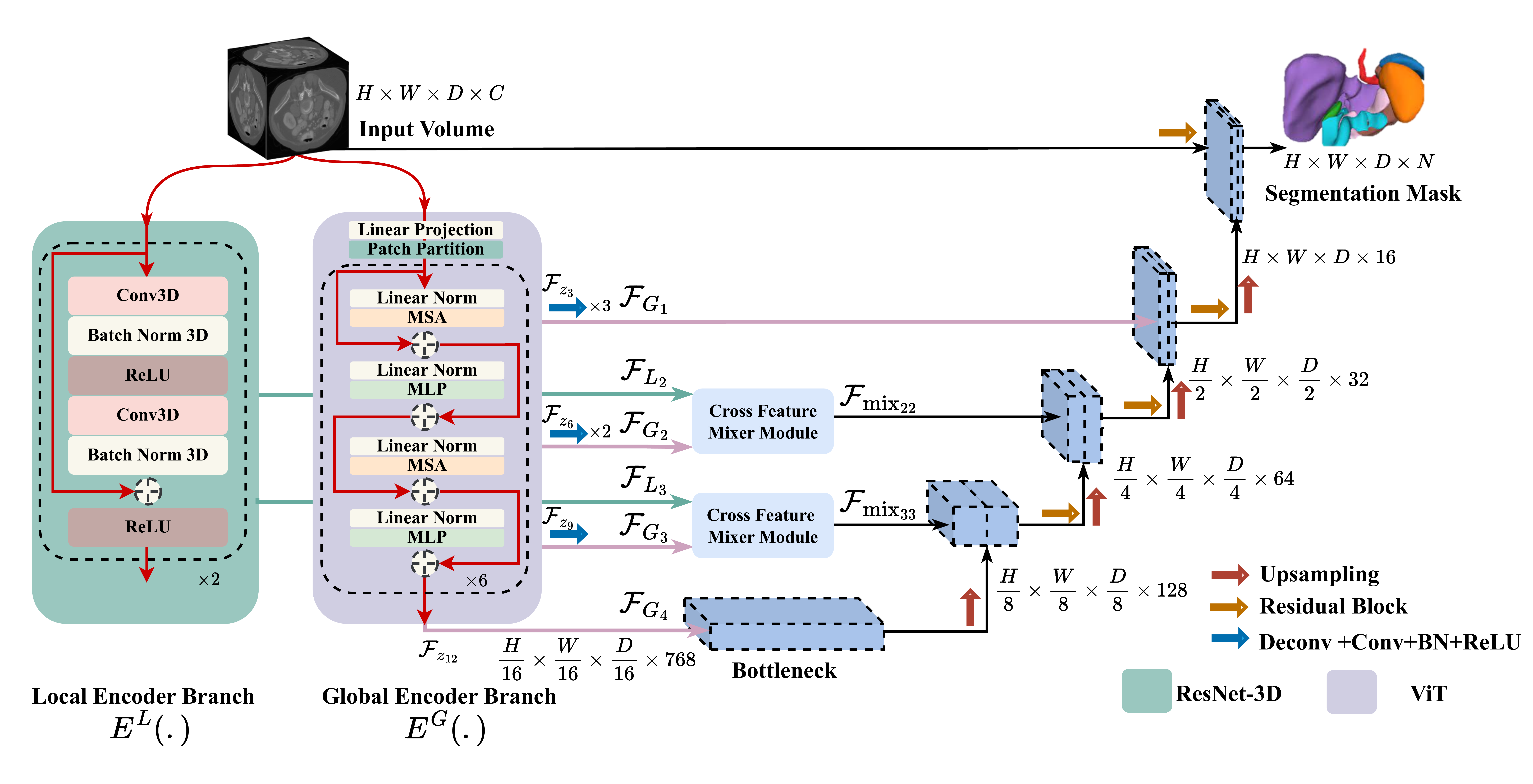}
\end{center}
\vspace{-2.5em}
\caption{\footnotesize An overview of the architecture Y-CT-Net designed for volumetric medical image segmentation, which takes a 3D input volume and passes it through two encoder branches. The local encoder branch consists of a ResNet3D, which learns local features using 3D convolution operations. The global encoder branch uses the backbone of a vision transformer to extract global features using a multi-head self-attention mechanism. Two stages of ResNet3D (Stage 2 and Stage 3) are used, with ResNet blocks of [4, 16]. The extracted features from both the local and global branches are fed into the \emph{Cross Feature Mixer Module (CFMM)}, which integrates local and global features to capture both local and global context information. The output of the CFMM and bottleneck module are then fed to a CNN-decoder via skip connections at multiple resolutions to predict the segmentation mask.}
\vspace{-1.0em}\label{fig:arch}
\vspace{-0.5em}
\end{figure*}

Given a volumetric image $\textbf{X} \in\R^{H \times W \times D \times C}$, we aim to predict the dense segmentation mask $\textbf{Y} \in\R^{H \times W \times D \times N}$ where $H$, $W$, $D$ represents the spatial dimensions, $C$ denotes channel dimension, and $N$ represents the number of predicted classes respectively. As our architecture consists of two parallel encoder branches, the volume $\textbf{X}$ is first prepossessed before passing through each encoder branch. For instance, input embedding, such as patch embedding for ViTs \cite{dosovitskiy2020image} is applied on volume $\textbf{X}$ before passing through $\textbf{E}^G(.)$ \vspace{-0.5em}
\begin{equation}
    \textbf{X}_G = \text{InputEmb}(\textbf{X}),\vspace{-0.5em}
\end{equation}
where $\textbf{X}_G \in\R^{N\times P^3 \times C \times K}$ denotes embedding tokens with patch resolution $P^3$,  sequence length $N$, and embedding dimension $K$. For input of $\textbf{E}^L(.)$, we apply appropriate operation on channel dimension of $\textbf{X}$ and form $\textbf{X}_L \in\R^{H \times W \times D \times 3C}$, for Synapse dataset $C = 1$. Then, each branch learns multi-scale local and global features maps that can be expressed as:\vspace{-0.5em}
\begin{equation}
\mathcal{F}_{L_i}=\textbf{E}^L(\textbf{X}_L) \hspace{2em}  ; \hspace{1em}i  \in  (1,2,3,..,M),\vspace{-0.5em}
\end{equation}
\begin{equation}
\mathcal{F}_{G_j}=\textbf{E}^G(\textbf{X}_G) \hspace{2em}  ; \hspace{1em}j  \in  (1,2,3,..,N),
\end{equation}
where $\mathcal{F}_{L_i}$ and $\mathcal{F}_{G_j}$ represent the features from $i_{\text{th}}$ and $j_{\text{th}}$ stage of local and global encoder branch respectively. These extracted feature maps are passed through the \emph{Cross Feature Mixer Module (CFMM)}, which has been designed carefully after conducting comprehensive ablation studies (See Table \ref{table:mixing_ablation_table}), to mix them based on their spatial dimensions. The mixing mechanism can be expressed as:\vspace{-0.5em}
\begin{equation}
\mathcal{F}_{\text{mix}_{ij}} =
\begin{cases}
\mathcal{F}_{L_i} \oplus \mathcal{F}_{G_j} & \text{if } C_{L_i} = C_{G_j}, \\
\text{Conv}(\mathcal{F}_{L_i}) \oplus \mathcal{F}_{G_j} & \text{if } C_{L_i} \neq C_{G_j}, \
\end{cases}\vspace{-0.5em}\label{eq:eq4}
\end{equation}

\noindent where $\oplus$ denotes element-wise addition, $C_{L_i}$ and $C_{G_j}$ are the channel dimension of local and global features of any stage, $\text{Conv}(\cdot)$ represents a convolutional layer that projects $\mathcal{F}_{L_i}$ to match the channel dimension of $\mathcal{F}_{G_j}$, and $\mathcal{F}_{\text{mix}_{ij}}$ represents the mixed feature map at any stage of local and global encoder branch. The Pytorch-like code of the feature mixing is shown in Algorithm \ref{alg:code}. The enriched features learned by \emph{Cross Feature Mixer Module (CFMM)} are used in decoding process to perform more precise and accurate segmentation.

\noindent \textbf{Instantiations of Y-CA-Net:} Y-CA-Net describes a general architecture that can be used to obtain different models by specifying the concrete design of encoder-decoder backbones. As shown in Fig. \ref{fig:concept-diagram} (a), if the local and global encoder branches are specified as ResNet-3D and ViT, respectively, and the decoder is specified as ConvNet, Y-CA-Net can then be transformed into a Y-CT-Net model.\vspace{-0.5em}

\subsection{Y-CT-Net}\label{sec:y-ct-net}\vspace{-0.5em}
In this section, we will explain one of models Y-CA-Net named Y-CT-Net which is carefully designed after extensive ablation studies. In the Y-CT-Net model, the local encoder branch is purely convolution based (\eg ResNet-3D) and the global encoder branch is purely attention based (\eg ViT, Swin), while the decoder is ConvNet based. The detail of each component will be explained in the following subsections.\vspace{-1em}
\subsubsection{Local Encoder Branch}\label{sec:localencoder}\vspace{- 0.5em}
~~~ As discussed earlier, any traditional 3D CNN can be used for our local encoder branch $\textbf{E}^L(.)$. For this purpose, we have used the ResNet-3D backbone as a local encoder for extracting local features. It has been noticed that the simultaneous use of two encoders increases the computational overhead. Therefore, we design our ResNet-3D in such a manner that it does not increase the computational costs by leveraging local information. To achieve this goal, we have eliminated some stages from ResNet-3D and increased the number of ResNet blocks in the early stages to maximize the learning of local information from the deeper layers after performing the extensive ablation studies which have been provided in Table \ref{table:blocks_ablation}. When the input image $\textbf{X}_L \in\R^{H \times W \times D \times 3C}$ is passed through the local encoder, it generates the feature maps denoted as ${ \mathcal{F}_{L}
=\textbf{E}^L(\textbf{X}_L)}$. The vanilla architecture of ResNet-3D has five stages and feature map of each stage is denoted as $\{\mathcal{F}_{L_{i}} | i  \in  (1,2,3,4,5)\}$. The first stage of ResNet-3D uses a larger kernel size of 7 $\times$ 7 to extract feature maps. In our case, the 3D input tensor passing through ResNet-3D consumes a large amount of computational cost. 
As a result, we relinquish this stage and keep the other stages that preserve more local feature details. 
The feature maps extracted from the earlier stages of the network retain more structural patterns and preserve more local information of neighborhood voxels than those extracted from the later stages. Consequently, we also eliminate the fourth and fifth stages while retaining the second and third stages from ResNet-3D. To utilize the advantage of these two stages, we increase the number of ResNet blocks, which effectively increases the capacity of the network to learn local features. 
We use the ResNet block of [4, 16] for stages 2 and 3 respectively. The feature maps learned by these stages i.e.
$\{ \mathcal{F}_{L_{i}} \in \mathbb{R}^{  \frac{H}{2^{i}} \times \frac{W}{2^{i}} \times \frac{D}{2^{i}} \times (32 \times 2^{i-1})}\}_{i=2}^{3}$, are passed through \emph{Cross Feature Mixer Module (CFMM)}, where it gets mixed with global features ($\mathcal{F}_{G_2}$ and $\mathcal{F}_{G_3}$) to capture the characteristics of convolution and self-attention and use these rich semantic features for dense prediction of target mask as shown in Fig. \ref{fig:arch}. \vspace{-1.25em}
\subsubsection{Global Encoder Branch}\vspace{-0.5em}
~~~ In our framework Y-CT-Net, any state-of-the-art transformer-based architecture can be used for the global encoder branch $\textbf{E}^G(.)$, such as a monolithic (ViT backbone) or non-monolithic (Swin backbone) architecture, where multi-head self-attention captures the pairwise relationship between patches. In our work, we use both transformer architecture as the global encoder, but here we describe vision transformer (ViT) as the global encoder branch which is responsible for learning global interactions. 

Specifically, in ViT, input image $\textbf{X}_G \in\R^{H \times W \times D \times C}$ is passed through the patch partition embedding layer where it is evenly divided into series of 3D non-overlapping patches with size $\frac{H}{P} \times \frac{W}{P} \times \frac{D}{P}$ each, and then projected into a $K$ dimensional space by linear projection, which remains constant throughout the transformer layers to yield feature maps denoted as ${ \mathcal{F}_{G}
=\textbf{E}^G(\textbf{X}_G)}$. After the patch embedding process, extracted patches are directly fed to successive transformer blocks, which comprises of multi-head self-attention (MSA) and multi-layer perceptron (MLP) alternately to extract the high-level feature maps denoted as $\{\mathcal{F}_{z_{i}} | i  \in (3,6,9,12)\}$ with their feature embedding dimension written as $\{ \mathcal{F}_{z_{i}}\in\mathbb{R}^{  \frac{H}{P} \times \frac{W}{P} \times \frac{D}{P} \times 768}\}_{i=3}^{12}$, where $P=16$ is dimension of patch size and $768$ is the embedding dimension. These feature maps are reshaped from embedding space into input space by applying consecutive $3\times 3 \times 3$ convolutional layers followed by normalization layers to get multi-scale features with new dimension written as
$\{\mathcal{F}_{G_{j}} \in \mathbb{R}^{  \frac{H}{2^{j}} \times \frac{W}{2^{j}} \times \frac{D}{2^{j}} \times (8 \times 2^{j})}\}_{j=1}^{3}$. For simplicity, we represent the features maps $\{\mathcal{F}_{z_{i}} | i \in (3,6,9,12)$ with new notation  $\{\mathcal{F}_{G_{j}} | j  \in (1,2,3,4)\}$. Based on the spatial dimension, feature maps from stages 2 and 3 (\ie $\mathcal{F}_{G_2}$ and $\mathcal{F}_{G_3}$), are passed through \emph{Cross Feature Mixer Module (CFMM)}, where they are mixed with local features to make dense predictions as shown in Fig. \ref{fig:arch}.\vspace{-1em}

\begin{table*}[!ht]
\caption{\footnotesize Quantitative results on multi-organ segmentation task using Synapse datasets compared with state-of-the-art methods. The green and red color cells represent the highest and the second highest scores respectively. Note that Aor: Aorta, Gal: Gallbladder, Lki: Left Kidney, Rki: Right Kidney, Liv: Liver, Pan: Pancreas, Spl: Spleen, Sto: Stomach are eight organ classes respectively.}\label{table:multi-organ}
\vspace{-0.5em}
\centering
\setlength{\tabcolsep}{19pt}
\scalebox{0.65}{
\begin{tabular}{l|c|c|c|c|c|c|c|c|c|c}
\toprule
\rowcolor{gray!50}
\textbf{Methods} & \textbf{Aor} & \textbf{Gal} & \textbf{Lki}  & \textbf{Rki} & \textbf{Liv} & \textbf{Pan}  & \textbf{Spl}  & \textbf{Sto} &\textbf{mDice} & \textbf{mHD95}\\
\midrule
R50 U-Net \cite{chen2021transunet}  & 0.877          & 0.636          & 0.806          & 0.781          & 0.937          & 0.569          & 0.858          & 0.741    & 0.746   & 0.368                \\
U-Net \cite{unet}      & 0.890          & \cellcolor{red!15}\textbf{0.697}          & 0.777          & 0.686          & 0.934          & 0.539          & 0.866          & 0.755       & 0.768  &  0.397   \\
R50 Att-UNet \cite{rattunet}  & 0.559          & 0.639          & 0.792          & 0.727          & 0.935          & 0.493          & 0.871          & 0.749 & 0.755   & 0.369               \\
Att-UNet \cite{attunet}    & 0.895          & 0.688          & 0.779          & 0.711          & 0.935          & 0.580          & 0.873          & 0.757    & 0.777   &  0.360           \\
ViT \cite{dosovitskiy2020image} + CUP \cite{chen2021transunet}  & 0.701 & 0.451 & 0.747 & 0.674 & 0.913 & 0.420 & 0.817 & 0.704&  0.678 & 0.361\\
R50-ViT \cite{dosovitskiy2020image} + CUP \cite{chen2021transunet} &  0.737 & 0.551 & 0.758 & 0.722 & 0.915 & 0.459 & 0.819 & 0.739 & 0.712 & 0.328 \\
TransUNet \cite{chen2021transunet}  & 0.872  & 0.631          & 0.818  & 0.770 & 0.940 & 0.558 & 0.850 & 0.756  & 0.774 & 0.316           \\
Swin-UNet \cite{cao2021swinunet}    & 0.854 & 0.665          & 0.832   & 0.796 & 0.943  & 0.565  & 0.906  & 0.766  & 0.791 &  0.215             \\
TransClaw U-Net \cite{chang2021transclaw} & 0.858  & 0.613      & 0.848  & 0.793 & 0.943 & 0.576  & 0.877   & 0.735 & 0.780 & 0.263    \\
LeVit-Unet-384 \cite{xu2021levit} & 0.873  & 0.622          & 0.846   & 0.802  & 0.931  & 0.590 & 0.888    & 0.727& 0.785 & 0.168                 \\
MISSFormer \cite{huang2021missformer}  & 0.869  & 0.686      & 0.852 & 0.820  & 0.944 & 0.656 & 0.919  & 0.808    & 0.819 & 0.182                   \\
MT-UNET \cite{mtunet}  & 0.879 & 0.650  & 0.814 & 0.773     & 0.930 & 0.594    & 0.877  & 0.768 & 0.786  & 0.265       \\
UNETR \cite{hatamizadeh2022unetr} & 0.899 & 
0.605  & 
0.856  & 
0.847  & 
0.945  & 
0.592  & 
0.878  & 
0.739  & 
0.795  & 
0.273      \\
Swin UNETR \cite{hatamizadeh2022swin}   & 
0.891  & 
0.696  & 
0.823  & 
0.790  & 
0.934  & 
0.686  & 
0.897  & 
0.682  & 
0.800  & 
0.462    \\
nnFormer \cite{zhou2021nnformer} & 
\cellcolor{green!15}\textbf{0.920}& 
\cellcolor{green!15}\textbf{{0.701}} & 
0.865 & 
0.862 & 
\cellcolor{green!15}\textbf{0.968} & 
\cellcolor{green!15}\textbf{0.833} & 
\cellcolor{red!15}\textbf{0.905} & 
\cellcolor{red!15}\textbf{0.868} & 
\cellcolor{red!15}\textbf{0.865} & 
\cellcolor{red!15}\textbf{0.116} \\
\midrule
Y-CT-Net-(ViT) & 
0.911 & 
0.610 & 
\cellcolor{red!15}\textbf{0.883} & 
0.848 & 
0.960 & 
0.701 & 
0.902 & 
0.780 & 
0.824 &       
0.195\\
Y-CT-Net-(Swin) & 
0.908 & 
0.660 & 
0.864 & 
\cellcolor{red!15}\textbf{0.865} & 
\cellcolor{red!15}\textbf{0.961} & 
0.722 & 
0.862 & 
0.747 &  
0.823 & 
0.440    \\
Y-CH-Net  & 
\cellcolor{red!15}\textbf{0.913} & 
0.567 & 
\cellcolor{green!15}\textbf{0.921} & 
\cellcolor{green!15}\textbf{0.940} & 
\cellcolor{green!15}\textbf{0.968} & 
\cellcolor{red!15}\textbf{0.816}   & 
\cellcolor{green!15}\textbf{0.956} &  
\cellcolor{green!15}\textbf{0.872} & 
\cellcolor{green!15}\textbf{0.870} & 
\cellcolor{green!15}\textbf{0.086}  \\
\bottomrule
\end{tabular}}
\vspace{-1.5em}
\end{table*}
\subsubsection{Cross Feature Mixer Module (CFMM)}\label{sec:CFMM}
\vspace{-0.5em}
~~~ In our framework, we introduce the \emph{Cross Feature Mixer Module (CFMM)}, to learn the rich semantic features by mixing the encoded features from both local and global branches which is mathematically explained by Eq. \ref{eq:eq4}. 
Specifically, as part of the mixing procedure, we select the feature maps based on their spatial dimensions. For this purpose, we select earlier features maps $ \mathcal{F}_{L_{2}} \in \mathbb{R}^{\frac{H}{4} \times \frac{W}{4} \times \frac{D}{4} \times 64}$, $\mathcal{F}_{L_{3}} \in \mathbb{R}^{\frac{H}{8} \times \frac{W}{8} \times \frac{D}{8} \times 128}$ from the local encoder and features of later layers $\mathcal{F}_{G_{2}} \in \mathbb{R}^{\frac{H}{4} \times \frac{W}{4} \times \frac{D}{4} \times 64}$, $ \mathcal{F}_{G_{3}} \in \mathbb{R}^{\frac{H}{8} \times \frac{W}{8} \times \frac{D}{8} \times 128}$ from the global encoder. As both types of features have similar spatial dimensions, so features with matching dimensions are mixed by simply element-wise addition (denoted by $\oplus$) in the following manner: \vspace{-0.5em}
\begin{equation}
  \mathcal{F}_{\text{mix}_{22}} = \mathcal{F}_{L_{2}} \oplus \mathcal{F}_{G_{2}} \in \mathbb{R}^{\frac{H}{4} \times \frac{W}{4} \times \frac{D}{4} \times 64},
  \vspace{-0.5em}
\end{equation}
\begin{equation}
   \mathcal{F}_{\text{mix}_{33}} = \mathcal{F}_{L_{3}} \oplus \mathcal{F}_{G_{3}} \in \mathbb{R}^{\frac{H}{8} \times \frac{W}{8} \times \frac{D}{8} \times 128}, 
\end{equation}

\noindent where $\mathcal{F}_{\text{mix}_{22}}$ and $\mathcal{F}_{\text{mix}_{33}}$ are the semantically enriched features which are used in the decoding process for the dense mask prediction.\vspace{-1em}

\subsubsection{Decoder}\vspace{-0.5em}
~~~ The flow of multi-scale feature maps from encoder branches to the CNN-based decoder is shown in Fig. \ref{fig:arch}. At bottleneck, we apply transpose convolutional layer to upsample the features. These features maps are then concatenated with enriched feature $\mathcal{F}_{\text{mix}_{33}}$ from \emph{Cross Feature Mixture Module (CFMM)}, and passed through the residual block consisting of two 3 $\times$ 3 $\times$ 3 convolutional layers with instance normalization. The output is then upsampled using deconvolutional layer. In the subsequent stage, the same process is repeated with enriched feature $\mathcal{F}_{\text{mix}_{22}}$, and first stage output $\mathcal{F}_{G_{1}}$. To calculate accurate probabilities scores for the volumetric segmentation mask, we concatenate the residual features of the input volume with upsampled features and passe them through 1 $\times$ 1 $\times$ 1 convolutional layer with softmax activation. The whole decoding process is shown in Fig. \ref{fig:arch}.


\subsection{Objective Function}
\vspace{-0.5em} 
The entire network Y-CT-Net is trained end-to-end with the joint objective function $\mathcal{L}(M_{gt},\hat{M}_{S})$, which includes the soft dice loss and cross entropy loss for the voxel-wise segmentation maps, which is defined as follows:
\vspace{-0.5em}
\begin{equation}
\begin{aligned}
\tiny
 \mathcal{L}(M_{gt},\hat{M}_{S}) = {}& 1 - \frac{2}{J}\sum_{j=1}^{J}\frac{\sum_{i=1}^{I}M_{gt_{i,j}}\hat{M}_{S_{i,j}}}{   \sum_{i=1}^{I} \hat{M}_{S_{i,j}}^2 + \sum_{i=1}^{I}M_{gt_{i,j}}^2} \\
 & - \frac{1}{I}\sum_{i=1}^{I}\sum_{j=1}^{J}M_{gt_{i,j}}\hat{M}_{S_{i,j}}
 \end{aligned}
\end{equation}

\noindent where $I$ and $J$ represent the number of voxels, and classes respectively. Here, $\hat{M}_{S_{i,j}}$, and $M_{gt_{i,j}}$, denotes the probability scores of the segmentation maps and one-hot encoded ground truth for the class $j$ at voxel $i$, respectively.\vspace{-0.5em}

\section{Experiments}\vspace{-0.5em}
In order to evaluate the effectiveness of our proposed method, we extensively conduct experiments on three publicly available datasets with different imaging modalities (CT, MRI): \textbf{(a)} multi-organ segmentation (Synapse dataset) \cite{landman2015miccai}, \textbf{(b)} brain tumor segmentation (BraTS-2021 Challenge dataset) \cite{Baid2021TheRB}, and \textbf{(c)} medical segmentation decathlon task segmentation (spleen dataset) \cite{antonelli2022medical}. More details on the datasets and training strategy have been provided in Appendix \ref{sec:dataset} and \ref{sec:implementation-details} respectively, while quantitative and visual results have been presented in the following subsections. We have highlighted the highest and second-highest scores in the results Sec. \ref{sec:quantitative-results} with different evaluation metrics.\vspace{-0.5em}

\begin{table*}[!ht]
	\centering
	\caption{ \footnotesize Quantitative results on brain tumor segmentation task using BraTS-2021 dataset for five-fold cross-validation using mean dice metric. ET, WT and TC denote Enhancing Tumor, Whole Tumor and Tumor Core respectively. The first and second highest mean dice scores of average of all classes across each fold are highlighted by using green and red color cells. All methods are run for 100 epochs and results are reported with inference overlap (\textit{io}) of 0.5.}
	\label{tab:brain-tumor-results}
 \vspace{-0.5em}
 \centering
\setlength{\tabcolsep}{10pt}
\scalebox{0.65}{
\begin{tabular}{l|c|c|c|c|c|c|c|c|c|c|c|c|c|c|c|c}
\toprule
   \rowcolor{gray!50}
	\textbf{Methods}& \multicolumn{4}{c|}{\textbf{Swin UNETR}\cite{hatamizadeh2022swin}} &
	\multicolumn{4}{c|}{\textbf{UNETR}\cite{hatamizadeh2022unetr}} &
	\multicolumn{4}{c|}{\textbf{SegResNet}\cite{segresnet}} &\multicolumn{4}{c}{\textbf{Y-CT-Net}}\\
 \midrule
Dice Score & TC & WT & ET& Avg. & TC & WT & ET & Avg.& TC & WT & ET & Avg. & TC & WT & ET & Avg.\\ \midrule
Fold 1 & 0.882 & 0.915 & 0.839& \cellcolor{red!15}\textbf{0.878} & 0.859 & 0.898 & 0.831& 0.862& 0.867&0.889&0.834&0.863& 0.900 & 0.915 & 0.860 & \cellcolor{green!20}\textbf{0.891} \\
Fold 2 & 0.890 & 0.920 & 0.880& \cellcolor{red!15}\textbf{0.896} & 0.851 & 0.904 & 0.859& 0.871& 0.874 & 0.893 & 0.854 & 0.873& 0.908 & 0.927 & 0.892 & \cellcolor{green!15}\textbf{0.909} \\
Fold 3 & 0.894 & 0.913 & 0.870& \cellcolor{red!15}\textbf{0.892} & 0.857 & 0.894 & 0.839& 0.863 & 0.874& 0.891& 0.864& 0.876& 0.899 & 0.918 & 0.875 & \cellcolor{green!15}\textbf{0.897} \\
Fold 4 & 0.891 & 0.914 & 0.864& \cellcolor{red!15}\textbf{0.889} & 0.865 & 0.891 & 0.843& 0.866&0.888 & 0.901 & 0.859 & 0.882&  0.905 & 0.911 & 0.874 & \cellcolor{green!15}\textbf{0.896} \\
Fold 5 & 0.894 & 0.920 & 0.864& \cellcolor{red!15}\textbf{0.892} & 0.841 & 0.901 & 0.841& 0.861 & 0.875& 0.894 & 0.845& 0.876& 0.912 & 0.924 & 0.879 & \cellcolor{green!15}\textbf{0.905} \\
\midrule
Avg. &0.890 &0.916 & 0.863&\cellcolor{red!15}\textbf{0.889} & 0.854& 0.897&0.842 &0.864 & 0.875&0.893 &0.851 & 0.874&0.904 &0.919 &0.876&\cellcolor{green!15}\textbf{0.899}
 \\
 \bottomrule
 \end{tabular}	}
\end{table*}

\begin{figure*}[!ht]

\begin{minipage}{\textwidth}
\centering
  \begin{minipage}{0.80\linewidth}
~~~\scalebox{0.1}{{\usebox{\Spleen}}} \footnotesize Spleen ~~~ \scalebox{0.1}{{\usebox{\RKID}}} \footnotesize R-Kidney ~~~ \scalebox{0.1}{{\usebox{\LKID}}} \footnotesize L-Kidney ~~~ \scalebox{0.1}{{\usebox{\GALL}}} \footnotesize Gallbladder ~~~ \scalebox{0.1}{{\usebox{\LIVER}}} \footnotesize Liver ~~~ \scalebox{0.1}{{\usebox{\STOMACH}}} \footnotesize Stomach ~~~ \scalebox{0.1}{{\usebox{\AORTA}}} \footnotesize Aorta ~~~ 
\scalebox{0.1}{{\usebox{\PANC}}} \footnotesize Pancreas
  \end{minipage}
\\
\vspace{0.2em}
\begin{minipage}{0.13\textwidth}
      \centering
    \includegraphics[height=2.2cm, width=\linewidth]{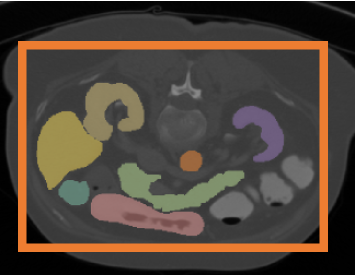}
  \end{minipage}
  \begin{minipage}{0.13\textwidth}
      \centering
   \includegraphics[height=2.2cm, width=\linewidth]{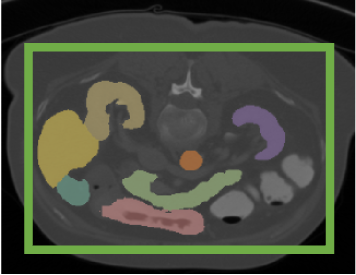}
  \end{minipage}
\begin{minipage}{0.13\textwidth}
      \centering
  \includegraphics[height=2.2cm, width=\linewidth]{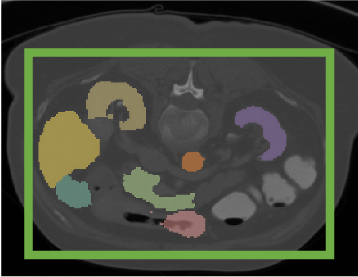}
  \end{minipage}
  \begin{minipage}{0.13\textwidth}
      \centering
    \includegraphics[height=2.2cm, width=\linewidth]{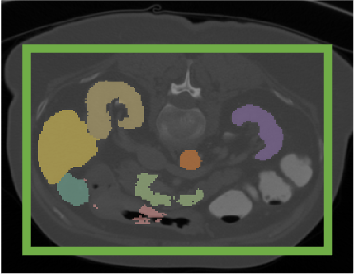}
  \end{minipage}
    \begin{minipage}{0.13\textwidth}
      \centering
   \includegraphics[height=2.2cm, width=\linewidth]{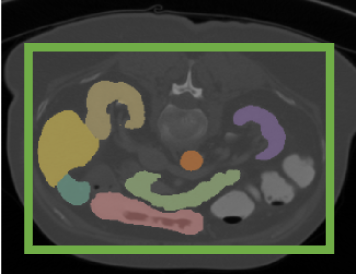}
  \end{minipage}
\begin{minipage}{0.13\textwidth}
      \centering
  \includegraphics[height=2.2cm, width=\linewidth]{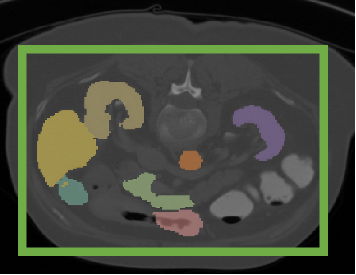}
  \end{minipage}
  \begin{minipage}{0.13\textwidth}
      \centering
    \includegraphics[height=2.2cm, width=\linewidth]{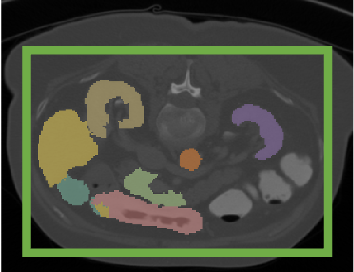}
  \end{minipage}
  \\
  \begin{minipage}{0.13\textwidth}
      \centering
   \includegraphics[height=2.2cm, width=\linewidth]{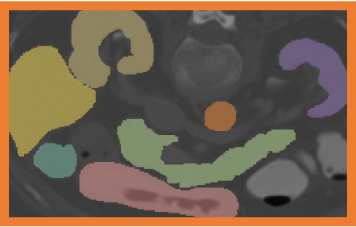}
   \footnotesize \textbf{LABEL}
  \end{minipage}
\begin{minipage}{0.13\textwidth}
      \centering
  \includegraphics[height=2.2cm, width=\linewidth]{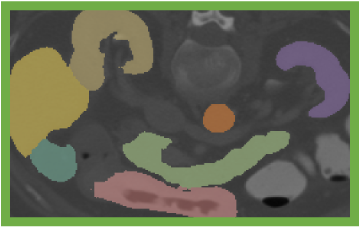}
  \footnotesize \textbf{nnFormer}
  \end{minipage}
  \begin{minipage}{0.13\textwidth}
      \centering
    \includegraphics[height=2.2cm, width=\linewidth]{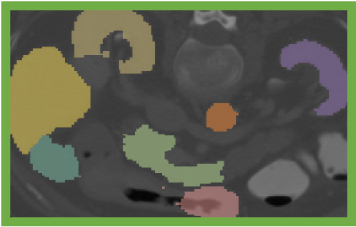}
    \footnotesize \textbf{Swin-UNETR}
  \end{minipage}
  \begin{minipage}{0.13\textwidth}
      \centering
   \includegraphics[height=2.2cm, width=\linewidth]{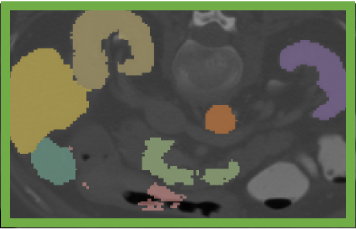}
   \footnotesize \textbf{UNETR}
  \end{minipage}
\begin{minipage}{0.13\textwidth}
      \centering
  \includegraphics[height=2.2cm, width=\linewidth]{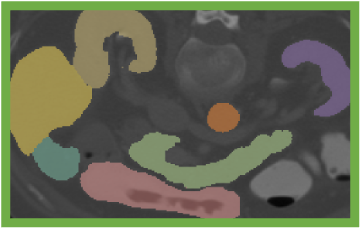}
  \footnotesize \textbf{Y-CH-Net}
  \end{minipage}
  \begin{minipage}{0.13\textwidth}
      \centering
    \includegraphics[height=2.2cm, width=\linewidth]{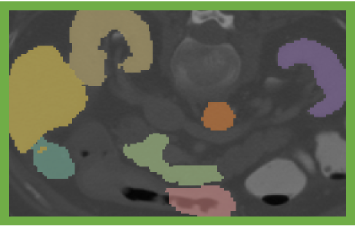}
    \footnotesize \textbf{Y-CT-Net(Swin)}
  \end{minipage}
  \begin{minipage}{0.13\textwidth}
      \centering
   \includegraphics[height=2.2cm, width=\linewidth]{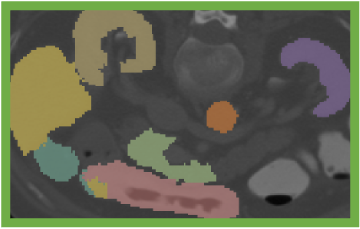}
   \footnotesize \textbf{Y-CT-Net(ViT)}
  \end{minipage}
\caption{\footnotesize Qualitative results on multi-organ segmentation. Visual results demonstrate that our method has not only performed accurate prediction but also preserve the organ information which is missed by each all three attention-based methods. For example, our Y-CH-Net is performing better than nnFormer, as boundary of stomach organ is being correctly predicted.}\label{fig:visualization-multi-organ}
\vspace{-0.5em}
\end{minipage}
\vspace{-1.5em}
\end{figure*}

\subsection{Quantitative Results}\label{sec:quantitative-results}\vspace{-0.5em}
We have used three attention-based method in our Y-CA-Net framework and evaluated it on three different segmentation datasets (mentioned in the Appendix \ref{sec:dataset}). The details of the results are mentioned as follows:\vspace{0.25em}

\noindent \textbf{Results of Multi-Organ Segmentation:} Table \ref{table:multi-organ} presents a comparison of the results obtained from ConvNets and transformer-based methods on the multi-organ segmentation task. Our proposed methods, Y-CT-Net and Y-CH-Net, show significant improvement in all evaluation metrics. When the hybrid \cite{zhou2021nnformer} or purely attention-based encoders architectures (ViT \cite{dosovitskiy2020image}, Swin \cite{liu2021swin}) are integrated with the local encoder branch, it results in an overall improvement margin of 3\%, 2.2\%, and 7.8\% in terms of mean Hausdorff distance \textit{(mHD95)}, respectively. Interestingly, for specific organs such as the left kidney, right kidney, spleen, and stomach, our method demonstrates impressive improvements both quantitatively and qualitatively, surpassing both hybrid and transformer-based methods.\vspace{0.25em}

\noindent\textbf{Results of Brain Tumor Segmentation:} The experimental results for the brain tumor segmentation task using five-fold cross-validation are presented in Table \ref{tab:brain-tumor-results}. Our proposed method, Y-CT-Net, outperforms Swin UNETR \cite{hatamizadeh2022swin} and SegResNet \cite{segresnet} methods that are solely designed for brain tumor segmentation by an average Dice score of 1\% and 2.5\%, respectively. As compared to other SOTA methods, Y-CT-Net is capable of effectively handling the delineation of Enhancing Tumor (ET), as demonstrated by both quantitative and qualitative results (See Table \ref{tab:brain-tumor-results} and Fig. \ref{fig:visualization-msd}).\vspace{0.25em}

\noindent \textbf{Results of MSD Task Segmentation:}
Table \ref{tab:msd}, demonstrates the effectiveness of our method. Although the increment is marginal, but our method has demonstrated improvements across all evaluation metrics.\vspace{-0.25em}

\subsection{Qualitative Results}
\begin{figure*}[!ht]

\begin{minipage}{0.5\textwidth}
\centering
\begin{minipage}{1.2\linewidth} 
\scalebox{0.1}{{\usebox{\TC}}} \footnotesize Whole Tumor (WT) ~ \scalebox{0.1}{{\usebox{\WT}}} \footnotesize Tumor Core (TC) ~ \scalebox{0.1}{{\usebox{\ET}}} \footnotesize Enhancing Tumor (ET)
\end{minipage}
\\
\vspace{0.2em}
\begin{minipage}{0.24\textwidth}
      \centering
    \includegraphics[height=2.4cm, width=\linewidth]{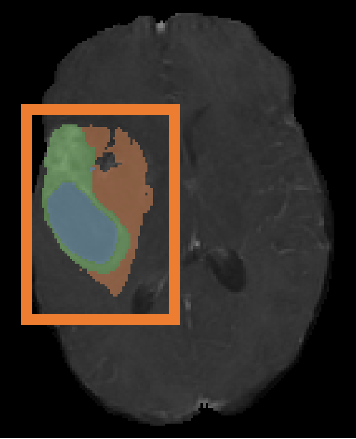}
  \end{minipage}
  \begin{minipage}{0.24\textwidth}
      \centering
   \includegraphics[height=2.4cm, width=\linewidth]{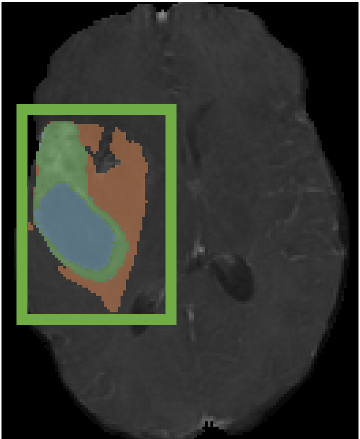}
  \end{minipage}
\begin{minipage}{0.24\textwidth}
      \centering
  \includegraphics[height=2.4cm, width=\linewidth]{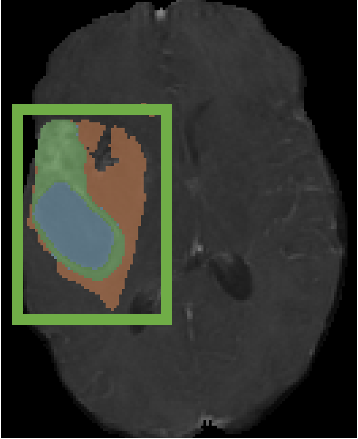}
  \end{minipage}
  \begin{minipage}{0.24\textwidth}
      \centering
    \includegraphics[height=2.4cm, width=\linewidth]{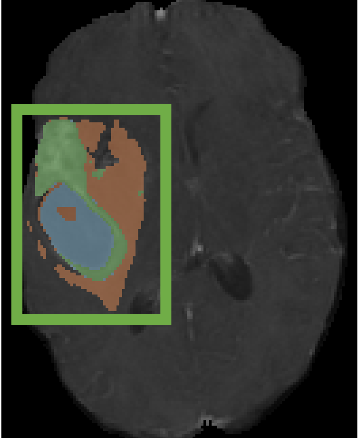}
  \end{minipage}
  \\
  \begin{minipage}{0.24\textwidth}
      \centering
   \includegraphics[height=2.4cm, width=\linewidth]{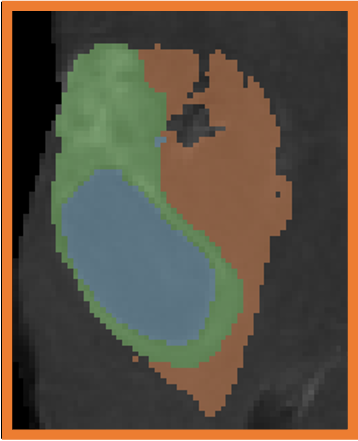}
   \footnotesize \textbf{LABEL}
  \end{minipage}
\begin{minipage}{0.24\textwidth}
      \centering
  \includegraphics[height=2.4cm, width=\linewidth]{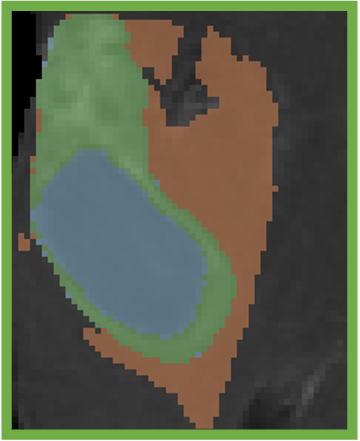}
  \footnotesize \textbf{Y-CT-Net}
  \end{minipage}
  \begin{minipage}{0.24\textwidth}
      \centering
    \includegraphics[height=2.4cm, width=\linewidth]{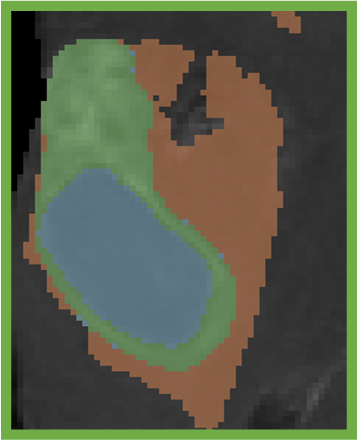}
    \footnotesize \textbf{Swin-UNETR}
  \end{minipage}
  \begin{minipage}{0.24\textwidth}
      \centering
   \includegraphics[height=2.4cm, width=\linewidth]{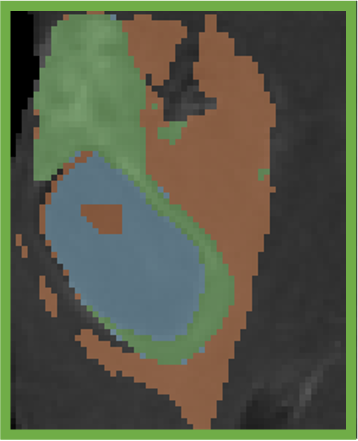}
   \footnotesize \textbf{UNETR}
  \end{minipage}
\caption{ \footnotesize{Qualitative results on brain tumor segmentation }}\label{fig:visualization-tumor} 
\end{minipage}
\hfill
\begin{minipage}{0.5\textwidth}
\centering
\begin{minipage}{\linewidth} 
\hspace{5em}\scalebox{0.1}{{\usebox{\MSD}}} \footnotesize Spleen 
\end{minipage}
\\
\vspace{0.2em}
\begin{minipage}{0.24\textwidth}
      \centering
    \includegraphics[height=2.4cm, width=\linewidth]{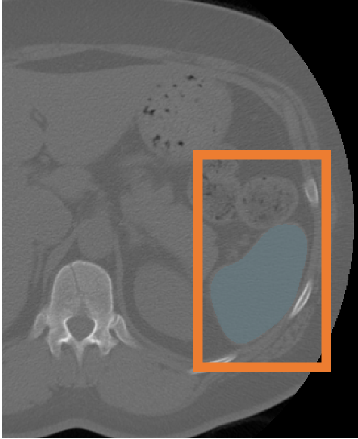}
  \end{minipage}
  \begin{minipage}{0.24\textwidth}
      \centering
   \includegraphics[height=2.4cm, width=\linewidth]{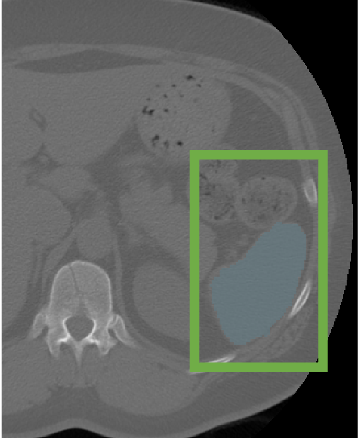}
  \end{minipage}
\begin{minipage}{0.24\textwidth}
      \centering
  \includegraphics[height=2.4cm, width=\linewidth]{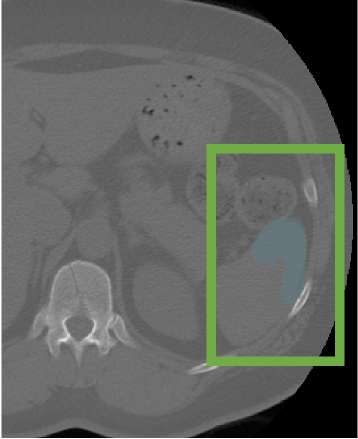}
  \end{minipage}
  \begin{minipage}{0.24\textwidth}
      \centering
    \includegraphics[height=2.4cm, width=\linewidth]{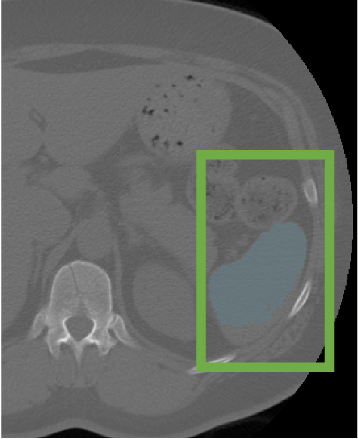}
  \end{minipage}
  \\
  \begin{minipage}{0.24\textwidth}
      \centering
   \includegraphics[height=2.4cm, width=\linewidth]{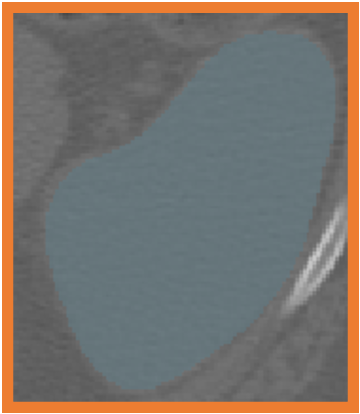}
   \footnotesize \textbf{LABEL}
  \end{minipage}
\begin{minipage}{0.24\textwidth}
      \centering
  \includegraphics[height=2.4cm, width=\linewidth]{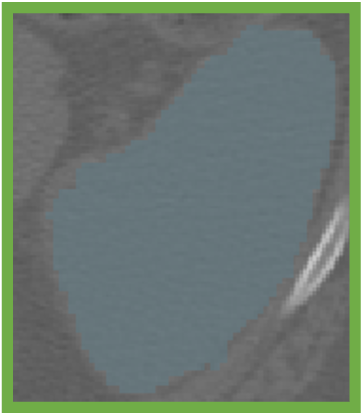}
  \footnotesize \textbf{Y-CT-Net}
  \end{minipage}
  \begin{minipage}{0.24\textwidth}
      \centering
    \includegraphics[height=2.4cm, width=\linewidth]{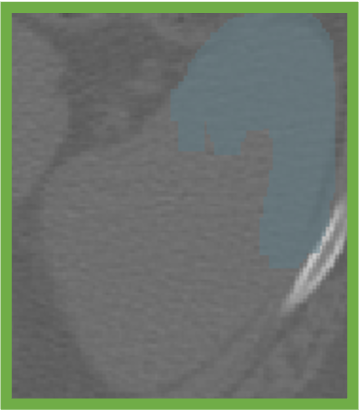}
    \footnotesize \textbf{Swin-UNETR}
  \end{minipage}
  \begin{minipage}{0.24\textwidth}
      \centering
   \includegraphics[height=2.4cm, width=\linewidth]{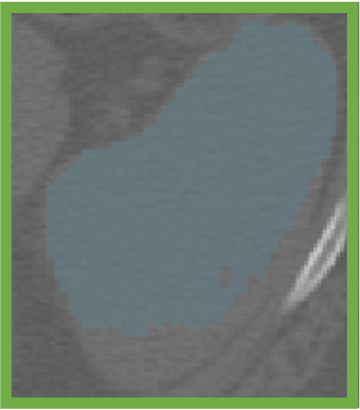}
   \footnotesize \textbf{UNETR}
  \end{minipage}
\caption{ \footnotesize{Qualitative results on MSD Spleen task segmentation}}\label{fig:visualization-msd} 
\end{minipage}
\vspace{-0.5em}
\end{figure*}

\vspace{-0.5em}
To intuitively demonstrate the effectiveness of our approach, we compare some qualitative results of our method with some recent state-of-the-art (SOTA) methods (nnFormer, Swin UNETR, and UNETR). Additional visual results have been provided in Appendix \ref{sec:add-results}. Fig. \ref{fig:visualization-multi-organ} shows the qualitative results for multi-organ segmentation. Our Y-CA-Net variants have better predicted multiple organ boundaries, including the stomach, pancreas, liver, and gallbladder, as compared to nnFormer, UNETR, and Swin UNETR. Specifically, our method has accurately preserved stomach boundaries, which have been missed by other methods. These results demonstrate that our Y-CA-Net variants have learned both local and global features, which are essential for volumetric segmentation. In Figures \ref{fig:visualization-tumor} and \ref{fig:visualization-msd}, we present the predicted segmentation masks for a brain tumor and MSD task segmentation for the spleen. Our method preserves fine-grained details while avoiding over-segmentation of the tumor regions. In the MSD task, our method has outperformed other SOTA methods by accurately predicting the spleen's boundary. 
\vspace{-0.5em}

\begin{table}[!ht]
\centering
\caption{\footnotesize Quantitative comparisons of the segmentation performance in spleen segmentation tasks of the MSD dataset.}
\label{tab:msd}
\vspace{-0.5em}
\setlength{\tabcolsep}{30pt}
\resizebox{0.43\textwidth}{!}{%
\begin{tabular}{l|c|c|c}
\toprule
\rowcolor{gray!50}
\multicolumn{1}{l}{\textbf{Task/Modality}}  
& \multicolumn{3}{c}{\textbf{Spleen Segmentation Task}} \\
\midrule
Metrics & mDice  & mPrec. & mSens. \\
\midrule
UNETR \cite{hatamizadeh2022unetr} &\cellcolor{red!15}\textbf{0.928} &\cellcolor{red!15}\textbf{0.923} &0.937  \\

Swin UNETR \cite{hatamizadeh2022swin}& 0.911 & 0.885 & \cellcolor{green!15}\textbf{0.955} \\

Y-CT-Net & \cellcolor{green!20}\textbf{0.934} & \cellcolor{green!20}\textbf{0.934} &\cellcolor{red!15}\textbf{0.940}  \\
\bottomrule
\end{tabular}}
\end{table}\vspace{-1em}

\vspace{-0.5em}

\begin{table*}[!ht]
     \begin{minipage}{.30\textwidth}
        \centering
\caption{\footnotesize{Ablation study of mixing the local and global features using different method.}\label{table:mixing_ablation_table}}
\vspace{-1em}
\small
\setlength{\tabcolsep}{31pt}
\scalebox{0.70}[0.70]{
\begin{tabular}{lc}
\toprule
\rowcolor{gray!50}
\textbf{Mixing Method}  & 
\textbf{Accuracy} \\
\midrule
Addition & \textbf{0.8392}\\
Averaging & 0.8008\\
Concatenation & 0.8390\\
Hadamard Product & 0.7930 \\
Self-attention & 0.8071\\
\bottomrule
\end{tabular}}
\end{minipage}
\hfill
\begin{minipage}{.30\textwidth}
\centering
\caption{\footnotesize Ablation study of using different feature pairs used in Cross Feature Mixer Module.}\label{table:position_ablation}
\vspace{-1em}
\small
\setlength{\tabcolsep}{4pt}
\setlength{\extrarowheight}{2pt}
\scalebox{0.70}[0.70]{
\begin{tabular}{lcc}
\toprule
\rowcolor{gray!50}
\textbf{Feature Pairs}  & \textbf{Spatial Dimension} &\textbf{Accuracy} \\
\midrule
($\mathcal{F}_{L_{2}}$, $\mathcal{F}_{G_{2}}$) & (24 $\times$ 24 $\times$ 24), (24 $\times$ 24 $\times$ 24) & \textbf{0.824}\\
($\mathcal{F}_{L_{3}}$, $\mathcal{F}_{G_{2}}$) & (12 $\times$ 12 $\times$ 12), (24 $\times$ 24 $\times$ 24) & 0.807\\
($\mathcal{F}_{L_{4}}$, $\mathcal{F}_{G_{2}}$) &  ($\phantom{0}$6 $\times$ $\phantom{0}$6 $\times$ $\phantom{0}$6), (24 $\times$ 24 $\times$ 24)& 0.798\\
($\mathcal{F}_{L_{5}}$, $\mathcal{F}_{G_{2}}$) &  ($\phantom{0}$3 $\times$ $\phantom{0}$3 $\times$ $\phantom{0}$3), (24 $\times$ 24 $\times$ 24) & 0.791\\
\bottomrule
\end{tabular}}
\end{minipage}
\hfill
\begin{minipage}{.30\textwidth}
\centering
\caption{\footnotesize Ablation study of using multiple Cross Feature Mixer Module.}\label{table:position_CFMM}
\vspace{-1em}
\small
\setlength{\tabcolsep}{3pt}
\setlength{\extrarowheight}{4.5pt}
\scalebox{0.70}[0.70]{
\begin{tabular}{lcc}
\toprule
\rowcolor{gray!50}
\textbf{Feature Pairs}  & \textbf{Spatial Dimension} &\textbf{Accuracy} \\
\midrule
($\mathcal{F}_{L_{2}}$, $\mathcal{F}_{G_{2}}$) & (24$^3$, 24$^3$) & 0.824\\
($\mathcal{F}_{L_{3}}$, $\mathcal{F}_{G_{3}}$) & (12$^3$, 12$^3$) & 0.823 \\
($\mathcal{F}_{L_{2}}$, $\mathcal{F}_{G_{2}}$)
($\mathcal{F}_{L_{3}}$, $\mathcal{F}_{G_{3}}$)&  (24$^3$, 24$^3$)(12$^3$, 12$^3$) & \textbf{0.839}\\

\bottomrule
\end{tabular}}
\end{minipage}
\vspace{-1.5em}
\end{table*}

\section{Ablative Analysis}\label{sec:ablation}\vspace{-0.5em}
To evaluate the effectiveness of our approach, we have performed several ablative studies. In all ablative analyses, we have used a global encoder branch with ViT backbone and used the Synapse dataset with limited samples to study the impact of these ablations on network performance in all our experiments. These studies are as follows:\vspace{-1em}\\

\noindent \textbf{Method of Mixing Features:} We conduct an ablation to study the impact of the performance of the network by using different methods of mixing local and global features. In Table \ref{table:mixing_ablation_table}, we present the quantitative results using the Synapse dataset. From Table \ref{table:mixing_ablation_table}, we observe that the performance of the network increases, when local and global features are mixed using the simple element-wise addition. These findings suggest that enriched features are learned by the network with simple addition operation without adding any complexity.\vspace{-1em}\\

\noindent \textbf{Position of Mixing Features:} For the better design of the \emph{Cross Feature Mixer Module (CFMM)}, we conduct an ablation study, in which we change the feature map pairs of both local and global branches used in the CFMM module, and study its effect on the network performance. From Table \ref{table:position_ablation}, we analyze that performance of the network increases, when the earlier stage feature maps of the local branch are mixed with feature maps of the global branch with a similar spatial dimension. This compelling study highlights the importance of local features and suggests that early stages of the local branch preserve more structural information and fine details which are significant for the dense prediction of segmentation mask.\vspace{-1em}\\

\noindent \textbf{Use of Multiple Cross Feature Mixer Module (CFMM):} Based on our previous ablative analysis reported in Table \ref{table:position_ablation}, we find out that the early stage feature maps from local encoder are more effective when combined with global encoder features of the similar spatial dimension. Based on these findings, we use the multiple \emph{Cross Feature Mixer Module (CFMM)} at multiple positions to combine feature pairs and study their impact on the network performance. The quantitative results in Table \ref{table:position_CFMM} demonstrate that using \emph{Cross Feature Mixer Module (CFMM)} at multiple positions, where the spatial dimension of both encoder branches matches increases the network performance.\vspace{-1em} \\

\noindent \textbf{Choice of Using Different CNN Backbones:} We perform an ablation study to determine which ConvNet backbone is most suitable for our local encoder branch. We observe that the ResNet backbone enhances the overall network performance as demonstrated in Table \ref{table:backbone_ablation}.\vspace{-1em}\\

\noindent \textbf{Ablation Study on Number of  Blocks:} We perform an ablation to investigate the effectiveness of our method by increasing the depth of stages of local encoder branch. We have found by increasing the number of Residual blocks in stages 2 and 3, leads to a significant improvement in network performance highlighted in Table \ref{table:blocks_ablation}.\vspace{-0.75em}

\begin{table}[!ht]
     \begin{minipage}{.22\textwidth}
        \centering
\caption{\scriptsize Ablation study of using different CNN backbone for local encoder branch.}
\label{table:backbone_ablation}
\vspace{-1em}
\small
\setlength{\tabcolsep}{14pt}
\setlength{\extrarowheight}{3.6pt}
\scalebox{0.70}[0.70]{
\begin{tabular}{lc}
\toprule
\rowcolor{gray!50}
\textbf{Local Encoder Network}  & 
\textbf{Accuracy} \\
\midrule
ResNet-3D \cite{resnet} & \textbf{0.839}\\
ResNext-3D \cite{resnext} & 0.801\\
ConvMixer-3D \cite{convmixer} & 0.786\\
ConvNext-3D \cite{convnext}& 0.771 \\
Non-local Network \cite{nonlocal}& 0.788\\
\bottomrule
\end{tabular}}
\end{minipage}
\hfill
\begin{minipage}{.2\textwidth}
\centering
\caption{\scriptsize Ablation study of increasing number of blocks of local encoder network for early stages.}\label{table:blocks_ablation}
\vspace{-1em}
\small
\setlength{\tabcolsep}{18pt}
\scalebox{0.70}[0.70]{
\begin{tabular}{lc}
\toprule
\rowcolor{gray!50}
\textbf{Blocks}  & \textbf{Accuracy} \\
\midrule
{[}2, 2{]} & 0.782 \\
{[}2, 4{]}  & 0.801 \\
{[}4, 4{]} & 0.839 \\
{[}4, 8{]} & 0.841 \\
{[}8, 8{]} & 0.831 \\
{[}4, 16{]} & \textbf{0.862} \\
{[}4, 24{]} & 0.854\\
\bottomrule

\end{tabular}}
\end{minipage}
\end{table}\vspace{-2em}

\section{Conclusion}\vspace{-0.5em}
In this paper, we introduce Y-CA-Net, a generic architecture with any encoder-decoder backbone for extracting local and global features for volumetric medical image segmentation. We introduce a simple yet effective \emph{Cross Feature Mixer Module (CFMM)} for learning enriched features which improves segmentation performance. Our findings suggest that utilizing both ConvNet and attention-based methods in combination is more effective for segmentation than using them individually. In our future work, we plan to further evaluate the performance of Y-CA-Net in diverse settings, such as self-supervised learning and transfer learning, to explore its potential for broader applications in medical image analysis.



\maketitle

\appendix 
\noindent \begin{huge} \textbf{Appendix} \vspace{4mm} \end{huge}\vspace{-0.25em}

In this section, we provide additional details regarding:
\begin{itemize}
    \item Dataset Details (Appendix \ref{sec:dataset})\vspace{-0.5em}
    \item Implementation Details (Appendix \ref{sec:implementation-details})\vspace{-0.5em}
    \item Additional Qualitative Results (Appendix \ref{sec:add-results})\vspace{-0.5em}
    \item Y-CA-Net-Variants (Appendix \ref{sec:variants})\vspace{-0.5em}
\end{itemize}

\section{Datasaet}\label{sec:dataset}

\noindent \textbf{Multi-Organ Segmentation:} In order to perform the volumetric segmentation for multi-organ tasks, we use the Synapse dataset. This dataset includes 30 subjects of abdominal CT scans. For a fair comparison, we also follow the same number of train-test splits as mentioned in these papers \cite{zhou2021nnformer, chen2021transunet}.  We evaluate the model performance with the dice score (DSC) and Hausdorff distance (HD95) on eight abdominal organs, which include the aorta, gallbladder, spleen, left kidney, right kidney, liver, pancreas, and stomach.\vspace{0.25em}

\noindent \textbf{Brain Tumor Segmentation:} The proposed framework is also evaluated for the brain tumor segmentation task using the BraTS-2021 Challenge dataset \cite{Baid2021TheRB}. The BraTS challenging dataset includes 1251 subjects, each with four 3D MRI modalities (T1w, T1gd, T2W, FLAIR). The BraTS datasets contain the annotations of three tumor-sub regions: Whole Tumor (WT), Tumor Core (TC), and Enhancing Tumor (TC). As the test ground truth labels are not available, we report the results using a five-fold cross-validation scheme with a ratio of 80/20 using dice metric.\vspace{0.25em}

\noindent \textbf{Medical Segmentation Decathlon (MSD) Task Segmentation:} MSD contains datasets for ten different segmentation tasks for different organs. From these databases, we report the results of the spleen task. To ensure fairness, we split the data as mentioned in these papers \cite{zhou2021nnformer, hatamizadeh2022unetr}, where the ratio of training-validation-test sets are 80\%, 15\% and 5\%.\vspace{-0.5em}

\subsubsection{Preprocessing}\vspace{-0.5em}
For all the tasks, we have used similar preprocessing techniques and applied same augmentations, which are described in these papers \cite{hatamizadeh2022unetr, tang2022self, hatamizadeh2022swin, zhou2021nnformer}.

\section{Implementation Details}\label{sec:implementation-details}
We have executed all our experiments using Pytorch \cite{paszke2019pytorch} and trained the model using a single NIVIDIA A100 SXM4 GPU. In our framework Y-CA-Net, we choose three recent attention-based methods \cite{hatamizadeh2022unetr, hatamizadeh2022swin, zhou2021nnformer} as global encoder branch, for integration with local encoder branch. We first reproduce the results using official implementations provided by the corresponding authors \cite{ hatamizadeh2022unetr, hatamizadeh2022swin, zhou2021nnformer}. We refrain from performing any modifications to the hyper-parameters of the chosen attention based methods while performing experiments on multi-organ segmentation. However, for brain tumor segmentation and MSD segmentation, we train all methods with the batch size of 1, using AdamW optimizer \cite{loshchilov2017decoupled} algorithm, with a learning rate of 0.0001 for 100, and 600 epochs respectively. In all experiments, the local encoder and global branch are initialized with random weights. Furthermore, all the images are re-sampled to the same target spacing and various data augmentation strategies; random rotation, flipping, scaling, and intensity shift to increase the data diversity, volume to avoid overfitting.

\section{Additional Qualitative Results}\label{sec:add-results}

\begin{figure}[!ht]
\begin{minipage}{0.5\textwidth}
\centering
  \begin{minipage}{0.24\textwidth}
      \centering
   \includegraphics[height=2.4cm, width=\linewidth]{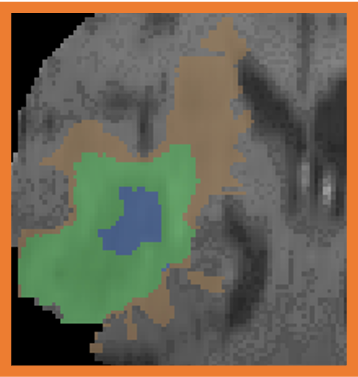}
  \end{minipage}
\begin{minipage}{0.24\textwidth}
      \centering
  \includegraphics[height=2.4cm, width=\linewidth]{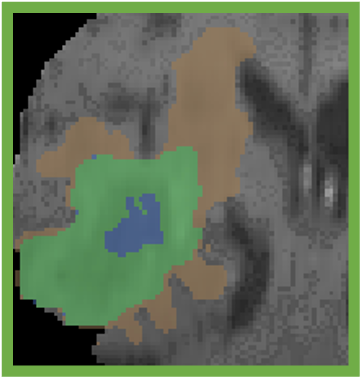}
  \end{minipage}
  \begin{minipage}{0.24\textwidth}
      \centering
    \includegraphics[height=2.4cm, width=\linewidth]{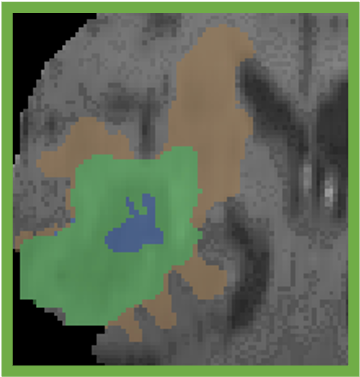}
  \end{minipage}
  \begin{minipage}{0.24\textwidth}
      \centering
   \includegraphics[height=2.4cm, width=\linewidth]{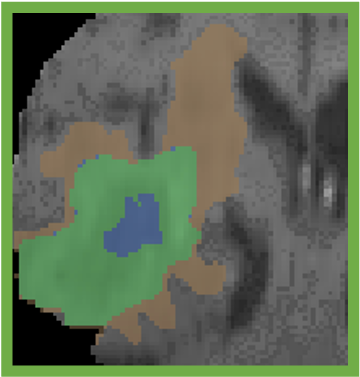}
  \end{minipage}
  \\
\begin{minipage}{0.24\textwidth}
      \centering
    \includegraphics[height=2.4cm, width=\linewidth]{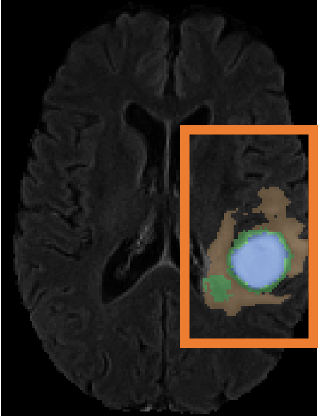}
  \end{minipage}
  \begin{minipage}{0.24\textwidth}
      \centering
   \includegraphics[height=2.4cm, width=\linewidth]{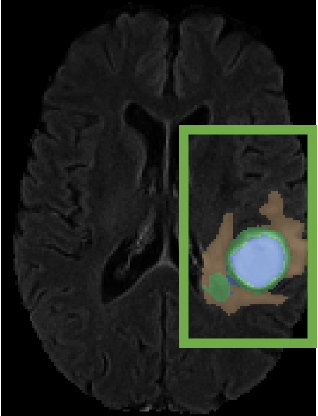}
  \end{minipage}
\begin{minipage}{0.24\textwidth}
      \centering
  \includegraphics[height=2.4cm, width=\linewidth]{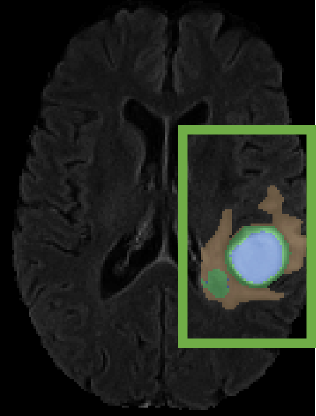}
  \end{minipage}
  \begin{minipage}{0.24\textwidth}
      \centering
    \includegraphics[height=2.4cm, width=\linewidth]{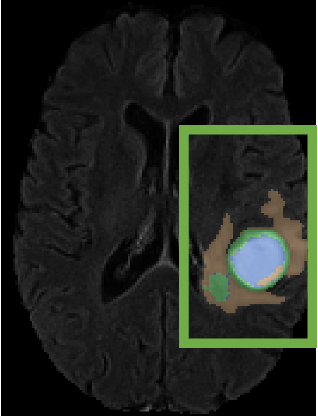}
  \end{minipage}
  \\
  \begin{minipage}{0.24\textwidth}
      \centering
   \includegraphics[height=2.4cm, width=\linewidth]{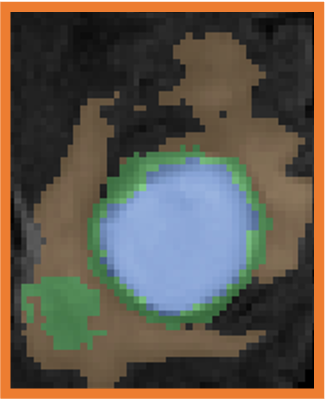}
     \footnotesize \textbf{LABEL}
  \end{minipage}
\begin{minipage}{0.24\textwidth}
      \centering
  \includegraphics[height=2.4cm, width=\linewidth]{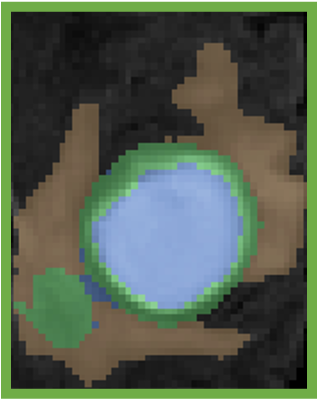}
  \footnotesize \textbf{Y-CT-Net}
  \end{minipage}
  \begin{minipage}{0.24\textwidth}
      \centering
    \includegraphics[height=2.4cm, width=\linewidth]{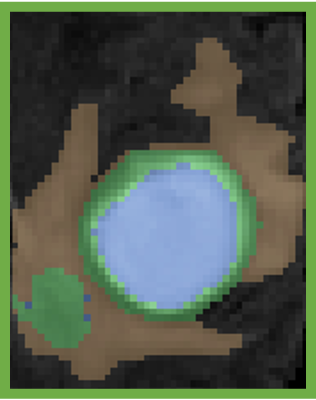}
    \footnotesize \textbf{Swin UNETR}
  \end{minipage}
  \begin{minipage}{0.24\textwidth}
      \centering
   \includegraphics[height=2.4cm, width=\linewidth]{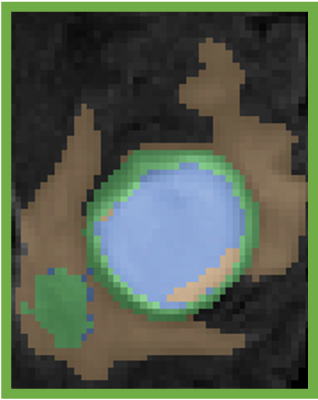}
   \footnotesize \textbf{UNETR}
  \end{minipage}
\end{minipage}
\caption{Qualitative results on brain tumor segmentation. Visual results demonstrates that our method has better predicted tumor regions.}\label{add-brain-results}\vspace{-0.5em}
\end{figure}

\begin{figure}[!ht]
\begin{minipage}{\textwidth} 
\scalebox{0.1}{{\usebox{\MSD}}} \footnotesize Spleen 
\end{minipage}
\begin{minipage}{0.5\textwidth}
\centering
\begin{minipage}{0.24\textwidth}
      \centering
    \includegraphics[height=2.4cm, width=\linewidth]{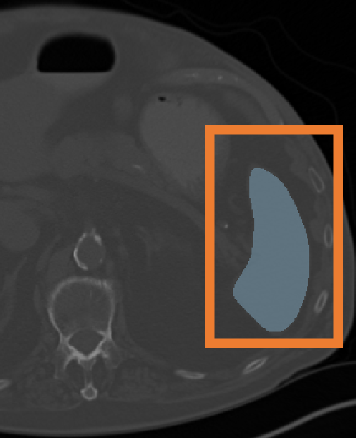}
  \end{minipage}
  \begin{minipage}{0.24\textwidth}
      \centering
   \includegraphics[height=2.4cm, width=\linewidth]{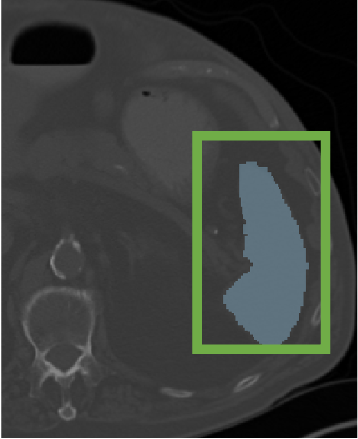}
  \end{minipage}
\begin{minipage}{0.24\textwidth}
      \centering
  \includegraphics[height=2.4cm, width=\linewidth]{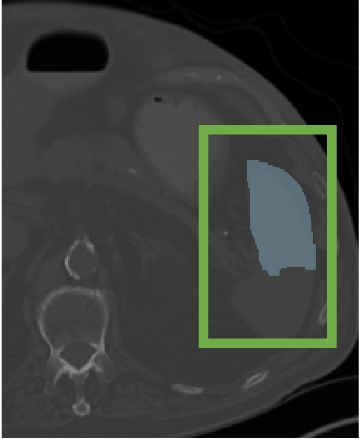}
  \end{minipage}
  \begin{minipage}{0.24\textwidth}
      \centering
    \includegraphics[height=2.4cm, width=\linewidth]{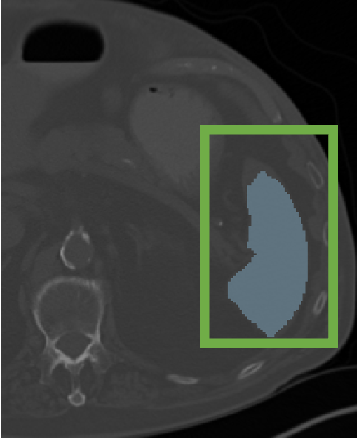}
  \end{minipage}
  \\
  \begin{minipage}{0.24\textwidth}
      \centering
   \includegraphics[height=2.4cm, width=\linewidth]{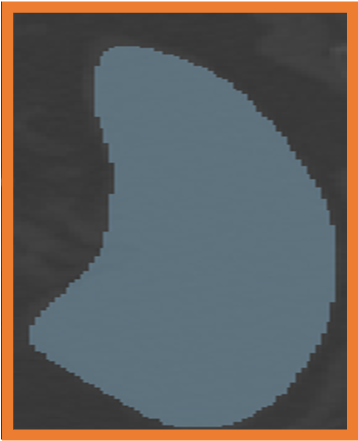}
  \end{minipage}
\begin{minipage}{0.24\textwidth}
      \centering
  \includegraphics[height=2.4cm, width=\linewidth]{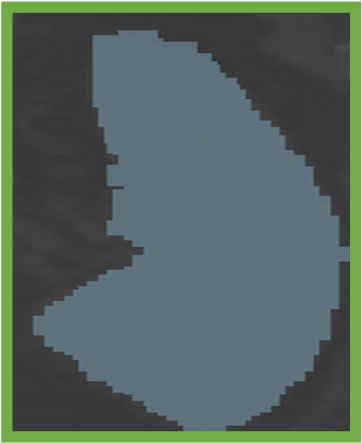}
  \end{minipage}
  \begin{minipage}{0.24\textwidth}
      \centering
    \includegraphics[height=2.4cm, width=\linewidth]{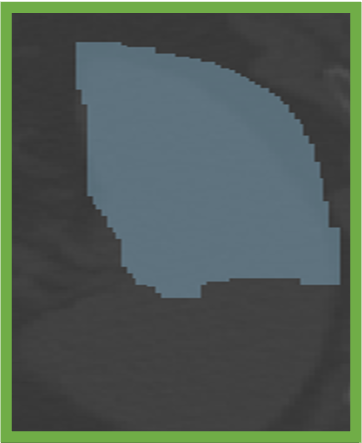}
  \end{minipage}
  \begin{minipage}{0.24\textwidth}
      \centering
   \includegraphics[height=2.4cm, width=\linewidth]{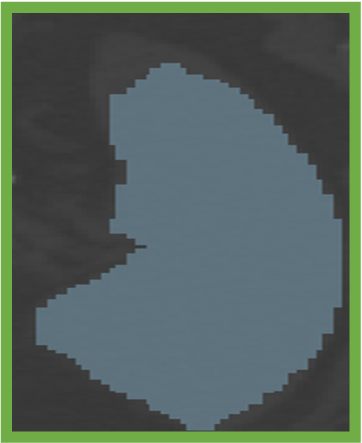}
  \end{minipage}
  \\
\begin{minipage}{0.24\textwidth}
      \centering
    \includegraphics[height=2.4cm, width=\linewidth]{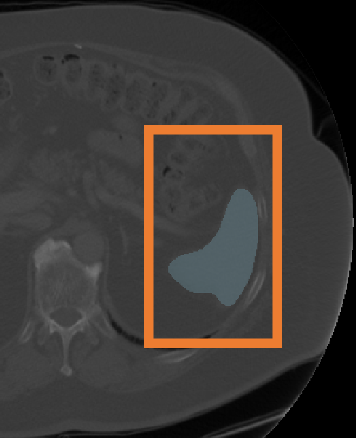}
  \end{minipage}
  \begin{minipage}{0.24\textwidth}
      \centering
   \includegraphics[height=2.4cm, width=\linewidth]{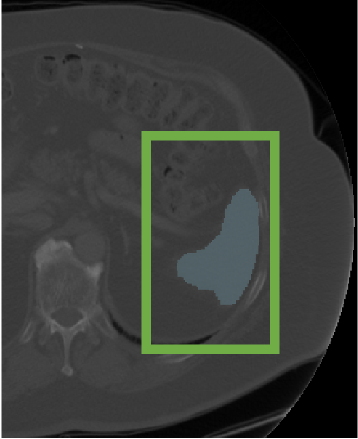}
  \end{minipage}
\begin{minipage}{0.24\textwidth}
      \centering
  \includegraphics[height=2.4cm, width=\linewidth]{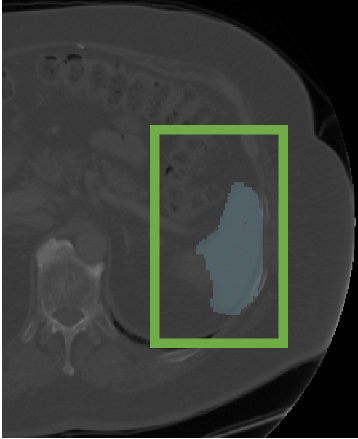}
  \end{minipage}
  \begin{minipage}{0.24\textwidth}
      \centering
    \includegraphics[height=2.4cm, width=\linewidth]{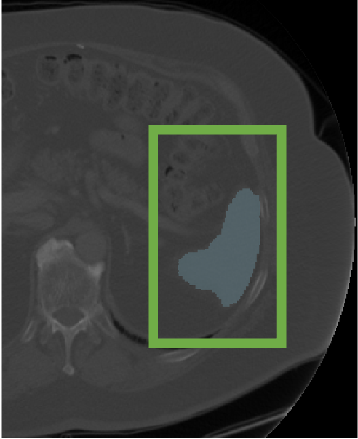}
  \end{minipage}
  \\
  \begin{minipage}{0.24\textwidth}
      \centering
   \includegraphics[height=2.4cm, width=\linewidth]{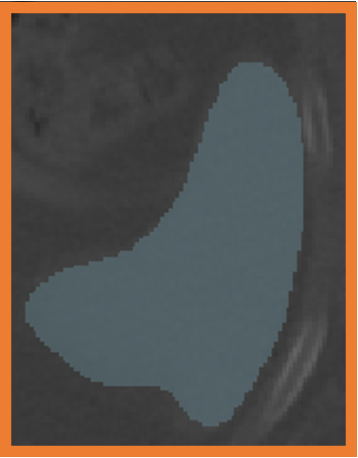}
    \footnotesize \textbf{LABEL}
  \end{minipage}
\begin{minipage}{0.24\textwidth}
      \centering
  \includegraphics[height=2.4cm, width=\linewidth]{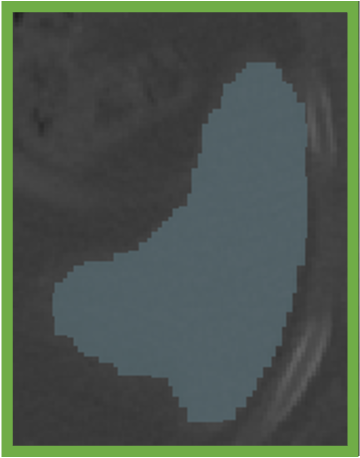}
   \footnotesize \textbf{Y-CT-Net}
  \end{minipage}
  \begin{minipage}{0.24\textwidth}
      \centering
    \includegraphics[height=2.4cm, width=\linewidth]{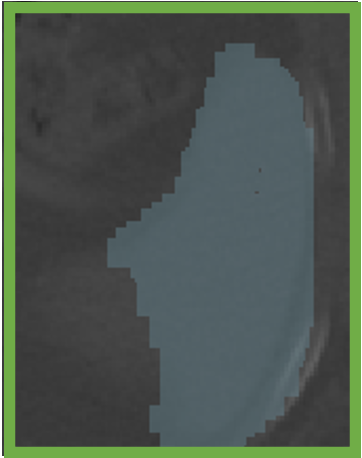}
    \footnotesize \textbf{Swin-UNETR}
  \end{minipage}
  \begin{minipage}{0.24\textwidth}
      \centering
   \includegraphics[height=2.4cm, width=\linewidth]{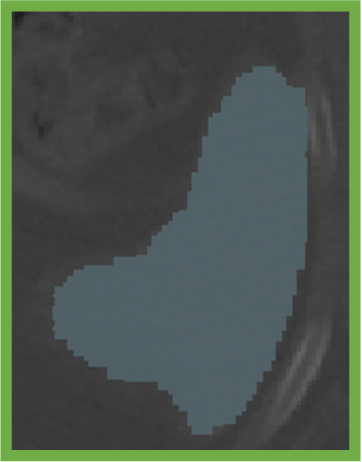}
   \footnotesize \textbf{UNETR}
  \end{minipage}
\end{minipage}
\caption{Qualitative results on MSD spleen task segmentation. Visual results demonstrates that our method has better predicted spleen organ without over-segmenting and preserving boundary information.}\label{add-spleen-results}
\end{figure}

We have provided the additional visual results for brain tumor, MSD spleen task and multi-organ segmentation in Figs. \ref{add-brain-results}, \ref{add-spleen-results}, and \ref{fig:visualization-multi-organ} respectively.

\section{Y-CA-Net Variants}\label{sec:variants}
In this section, we will explain two other variants of Y-CA-Nets, namely Y-CT-Net and Y-CH-Net. The detail of one variant Y-CA-Net which is actually Y-CT-Net with ViT backbone has already explained in the paper.\vspace{0.5em}

\noindent \textbf{Y-CT-Net:} In the Y-CT-Net model, the local encoder branch is the same as explained in the paper, and the global encoder branch uses the swin backbone. Feature maps with similar spatial dimensions are mixed in the \emph{Cross Feature Mixture Module (CFMM)}. In this variant, feature maps from the local encoder branch $\mathcal{F}_{L_2}$ and $\mathcal{F}_{L_3}$ are mixed with global encoder branch features $\mathcal{F}_{G_1}$ and $\mathcal{F}_{G_2}$, and these enriched features are used in the ConvNet-based decoder for dense prediction of segmentation masks.\vspace{0.5em}

\noindent \textbf{Y-CH-Net:} In the Y-CH-Net model, the local encoder branch is purely convolution-based, as explained in the paper. The global encoder branch is hybrid-attention-based (nnFormer \cite{zhou2021nnformer}), and the decoder is also hybrid-attention-based. Feature maps from the local encoder branch $\mathcal{F}_{L_2}$ and $\mathcal{F}_{L_3}$ are mixed with the global encoder branch (hybrid-attention-based) features of the second $\mathcal{F}_{G_2}$ and third stages $\mathcal{F}_{G_3}$, respectively in the CFMM module. These enriched features are then passed through a hybrid-attention-based decoder for dense prediction of segmentation masks.\vspace{0.5em}

In summary, Y-CT-Net and Y-CH-Net are two variants of Y-CA-Nets that use different combinations of encoder decoder backbones.  

\begin{figure}[!ht]
\begin{minipage}{\textwidth} 
\scalebox{0.1}{{\usebox{\TC}}} \footnotesize Whole Tumor (WT) ~ \scalebox{0.1}{{\usebox{\WT}}} \footnotesize Tumor Core (TC) ~ \scalebox{0.1}{{\usebox{\ET}}} \footnotesize Enhancing Tumor (ET)
\end{minipage}
\begin{minipage}{0.5\textwidth}
\centering
\begin{minipage}{0.24\textwidth}
      \centering
    \includegraphics[height=2.4cm, width=\linewidth]{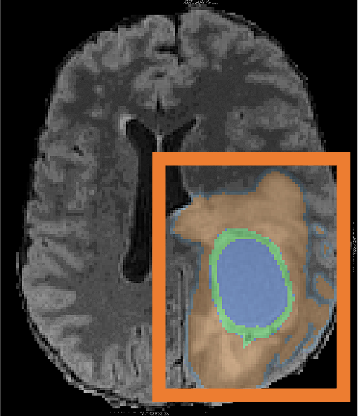}
  \end{minipage}
  \begin{minipage}{0.24\textwidth}
      \centering
   \includegraphics[height=2.4cm, width=\linewidth]{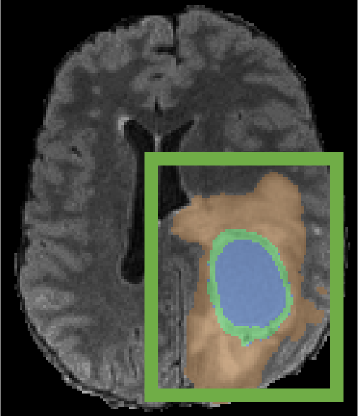}
  \end{minipage}
\begin{minipage}{0.24\textwidth}
      \centering
  \includegraphics[height=2.4cm, width=\linewidth]{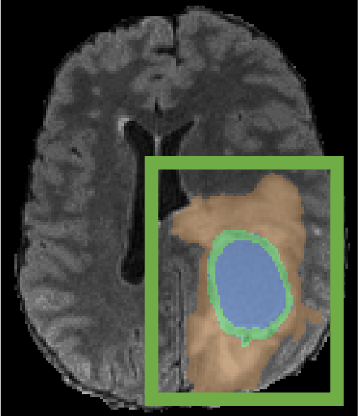}
  \end{minipage}
  \begin{minipage}{0.24\textwidth}
      \centering
    \includegraphics[height=2.4cm, width=\linewidth]{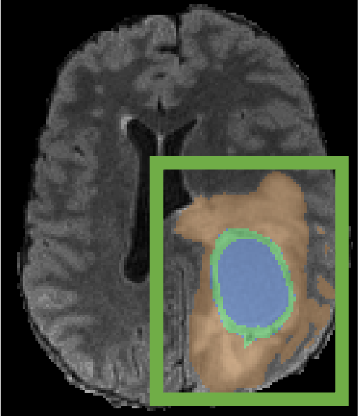}
  \end{minipage}
  \\
  \begin{minipage}{0.24\textwidth}
      \centering
   \includegraphics[height=2.4cm, width=\linewidth]{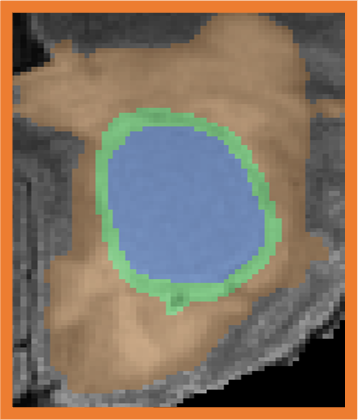}
  \end{minipage}
\begin{minipage}{0.24\textwidth}
      \centering
  \includegraphics[height=2.4cm, width=\linewidth]{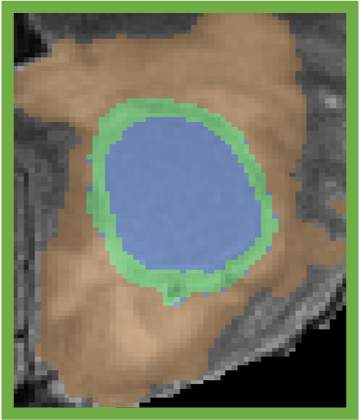}
  \end{minipage}
  \begin{minipage}{0.24\textwidth}
      \centering
    \includegraphics[height=2.4cm, width=\linewidth]{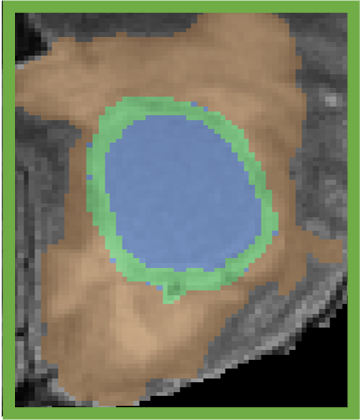}
  \end{minipage}
  \begin{minipage}{0.24\textwidth}
      \centering
   \includegraphics[height=2.4cm, width=\linewidth]{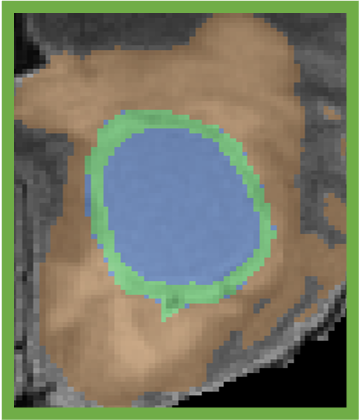}
  \end{minipage}
\begin{minipage}{0.24\textwidth}
      \centering
    \includegraphics[height=2.4cm, width=\linewidth]{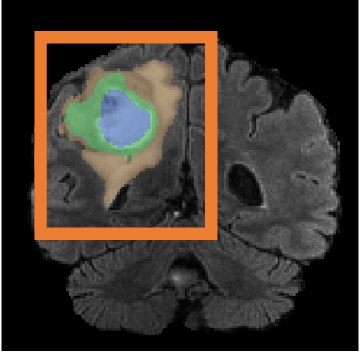}
  \end{minipage}
  \begin{minipage}{0.24\textwidth}
      \centering
   \includegraphics[height=2.4cm, width=\linewidth]{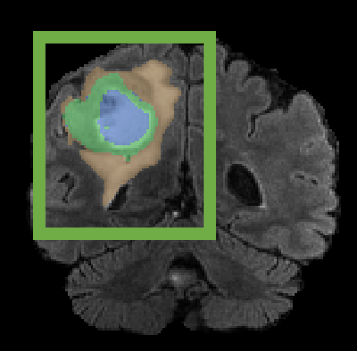}
  \end{minipage}
\begin{minipage}{0.24\textwidth}
      \centering
  \includegraphics[height=2.4cm, width=\linewidth]{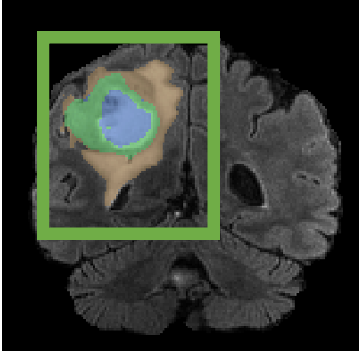}
  \end{minipage}
  \begin{minipage}{0.24\textwidth}
      \centering
    \includegraphics[height=2.4cm, width=\linewidth]{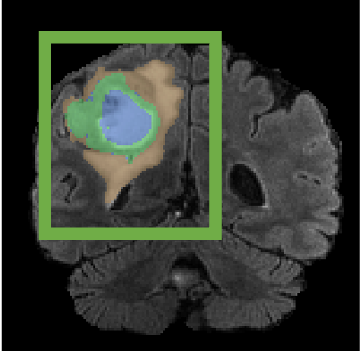}
  \end{minipage}
  \\
  \begin{minipage}{0.24\textwidth}
      \centering
   \includegraphics[height=2.4cm, width=\linewidth]{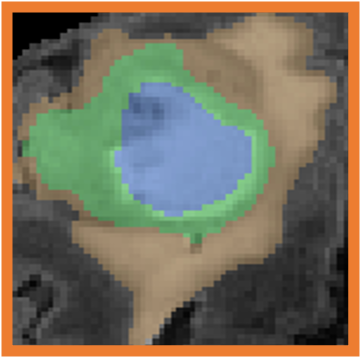}
  \end{minipage}
\begin{minipage}{0.24\textwidth}
      \centering
  \includegraphics[height=2.4cm, width=\linewidth]{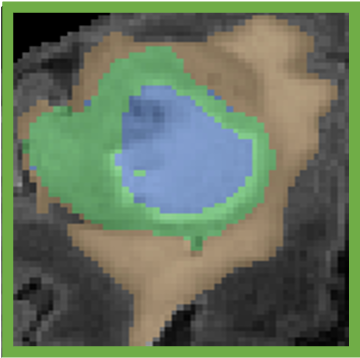}
  \end{minipage}
  \begin{minipage}{0.24\textwidth}
      \centering
    \includegraphics[height=2.4cm, width=\linewidth]{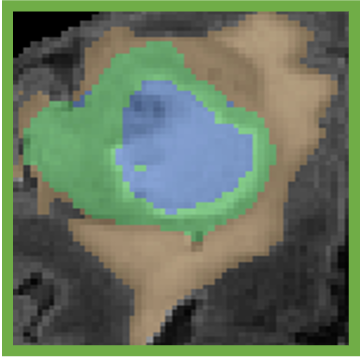}
  \end{minipage}
  \begin{minipage}{0.24\textwidth}
      \centering
   \includegraphics[height=2.4cm, width=\linewidth]{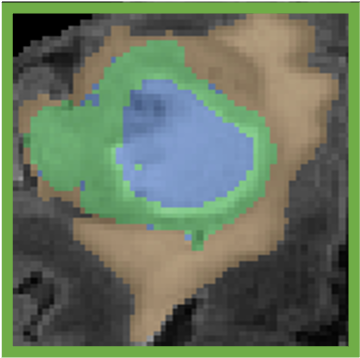}
  \end{minipage}
  \begin{minipage}{0.24\textwidth}
      \centering
    \includegraphics[height=2.4cm, width=\linewidth]{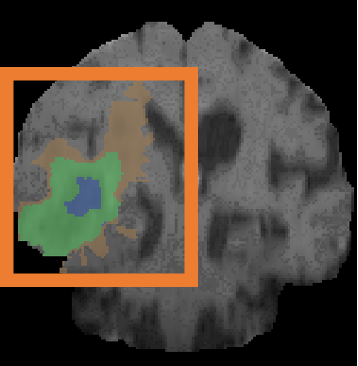}
  \end{minipage}
  \begin{minipage}{0.24\textwidth}
      \centering
   \includegraphics[height=2.4cm, width=\linewidth]{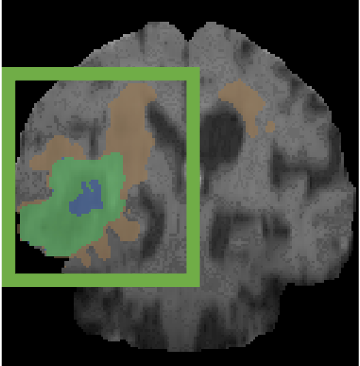}
  \end{minipage}
\begin{minipage}{0.24\textwidth}
      \centering
  \includegraphics[height=2.4cm, width=\linewidth]{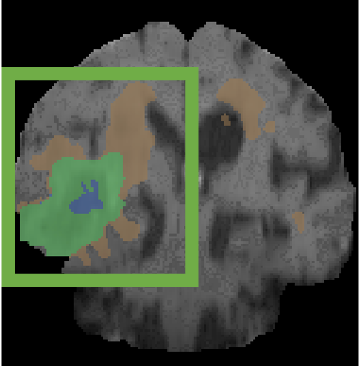}
  \end{minipage}
  \begin{minipage}{0.24\textwidth}
      \centering
    \includegraphics[height=2.4cm, width=\linewidth]{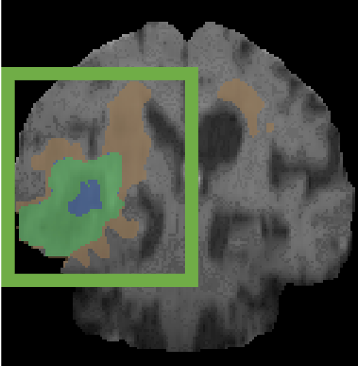}
  \end{minipage}
\end{minipage}
\end{figure}

\newpage

\begin{figure*}[!ht]

\begin{minipage}{\textwidth}
\centering
  \begin{minipage}{0.80\linewidth}
~~~\scalebox{0.1}{{\usebox{\Spleen}}} \footnotesize Spleen ~~~ \scalebox{0.1}{{\usebox{\RKID}}} \footnotesize R-Kidney ~~~ \scalebox{0.1}{{\usebox{\LKID}}} \footnotesize L-Kidney ~~~ \scalebox{0.1}{{\usebox{\GALL}}} \footnotesize Gallbladder ~~~ \scalebox{0.1}{{\usebox{\LIVER}}} \footnotesize Liver ~~~ \scalebox{0.1}{{\usebox{\STOMACH}}} \footnotesize Stomach ~~~ \scalebox{0.1}{{\usebox{\AORTA}}} \footnotesize Aorta ~~~ 
\scalebox{0.1}{{\usebox{\PANC}}} \footnotesize Pancreas
  \end{minipage}
\\
\vspace{0.2em}
\begin{minipage}{0.13\textwidth}
      \centering
    \includegraphics[height=2.2cm, width=\linewidth]{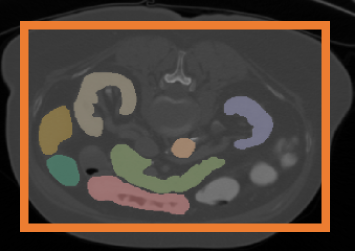}
  \end{minipage}
  \begin{minipage}{0.13\textwidth}
      \centering
   \includegraphics[height=2.2cm, width=\linewidth]{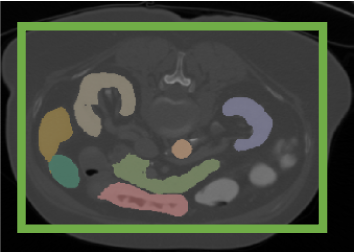}
  \end{minipage}
\begin{minipage}{0.13\textwidth}
      \centering
  \includegraphics[height=2.2cm, width=\linewidth]{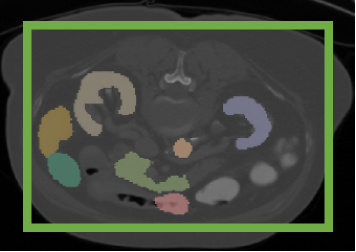}
  \end{minipage}
  \begin{minipage}{0.13\textwidth}
      \centering
    \includegraphics[height=2.2cm, width=\linewidth]{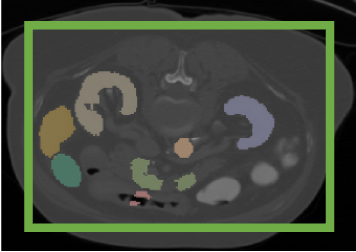}
  \end{minipage}
    \begin{minipage}{0.13\textwidth}
      \centering
   \includegraphics[height=2.2cm, width=\linewidth]{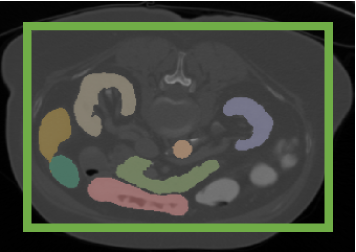}
  \end{minipage}
\begin{minipage}{0.13\textwidth}
      \centering
  \includegraphics[height=2.2cm, width=\linewidth]{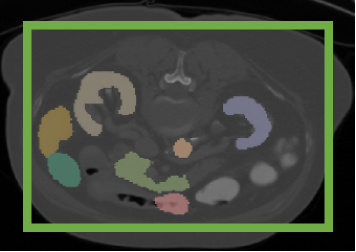}
  \end{minipage}
  \begin{minipage}{0.13\textwidth}
      \centering
    \includegraphics[height=2.2cm, width=\linewidth]{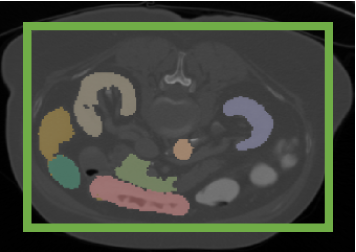}
  \end{minipage}
  \\
  \begin{minipage}{0.13\textwidth}
      \centering
   \includegraphics[height=2.2cm, width=\linewidth]{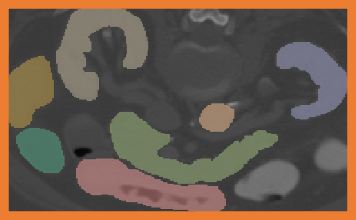}
  \end{minipage}
\begin{minipage}{0.13\textwidth}
      \centering
  \includegraphics[height=2.2cm, width=\linewidth]{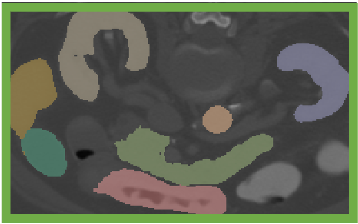}
  \end{minipage}
  \begin{minipage}{0.13\textwidth}
      \centering
    \includegraphics[height=2.2cm, width=\linewidth]{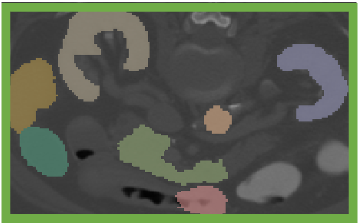}
  \end{minipage}
  \begin{minipage}{0.13\textwidth}
      \centering
   \includegraphics[height=2.2cm, width=\linewidth]{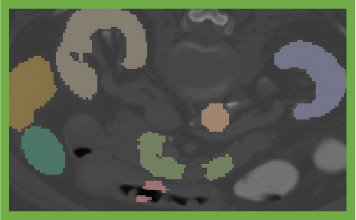}
  \end{minipage}
\begin{minipage}{0.13\textwidth}
      \centering
  \includegraphics[height=2.2cm, width=\linewidth]{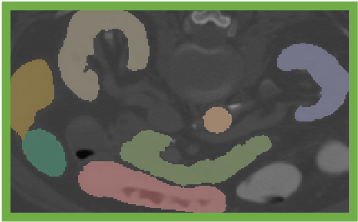}
  \end{minipage}
  \begin{minipage}{0.13\textwidth}
      \centering
    \includegraphics[height=2.2cm, width=\linewidth]{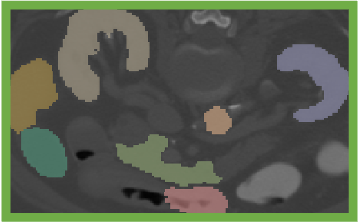}
  \end{minipage}
  \begin{minipage}{0.13\textwidth}
      \centering
   \includegraphics[height=2.2cm, width=\linewidth]{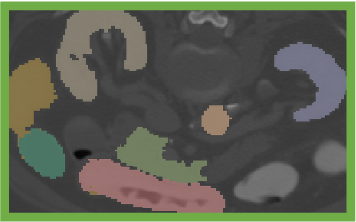}
  \end{minipage}
  \\
  \begin{minipage}{0.13\textwidth}
      \centering
   \includegraphics[height=2.2cm, width=\linewidth]{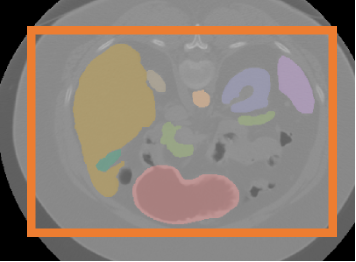}
  \end{minipage}
\begin{minipage}{0.13\textwidth}
      \centering
  \includegraphics[height=2.2cm, width=\linewidth]{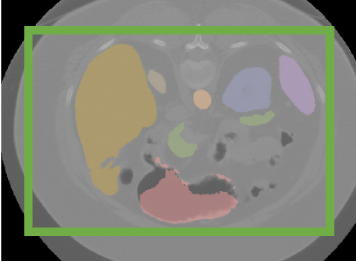}
  \end{minipage}
  \begin{minipage}{0.13\textwidth}
      \centering
    \includegraphics[height=2.2cm, width=\linewidth]{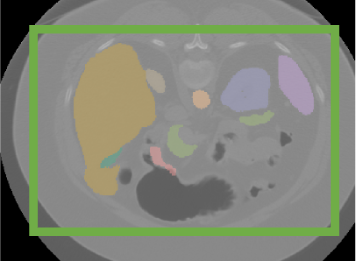}
  \end{minipage}
  \begin{minipage}{0.13\textwidth}
      \centering
   \includegraphics[height=2.2cm, width=\linewidth]{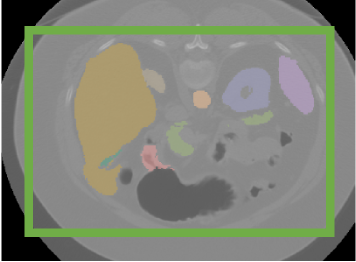}
  \end{minipage}
\begin{minipage}{0.13\textwidth}
      \centering
  \includegraphics[height=2.2cm, width=\linewidth]{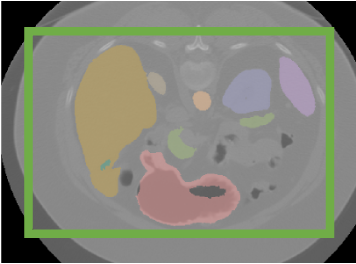}
  \end{minipage}
  \begin{minipage}{0.13\textwidth}
      \centering
    \includegraphics[height=2.2cm, width=\linewidth]{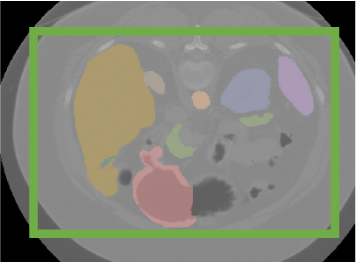}

  \end{minipage}
  \begin{minipage}{0.13\textwidth}
      \centering
   \includegraphics[height=2.2cm, width=\linewidth]{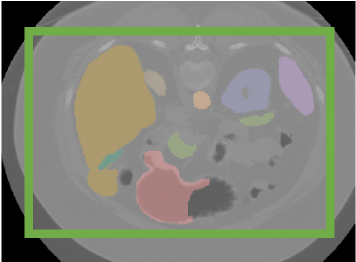}
  \end{minipage}
  \\
  \begin{minipage}{0.13\textwidth}
      \centering
   \includegraphics[height=2.2cm, width=\linewidth]{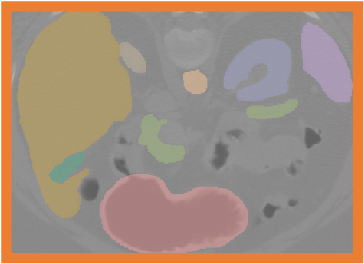}
   \footnotesize \textbf{LABEL}
  \end{minipage}
\begin{minipage}{0.13\textwidth}
      \centering
  \includegraphics[height=2.2cm, width=\linewidth]{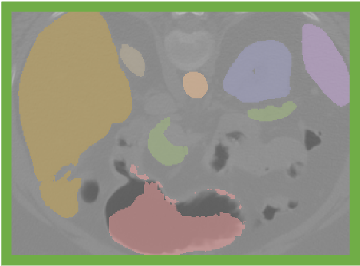}
  \footnotesize \textbf{nnFormer}
  \end{minipage}
  \begin{minipage}{0.13\textwidth}
      \centering
    \includegraphics[height=2.2cm, width=\linewidth]{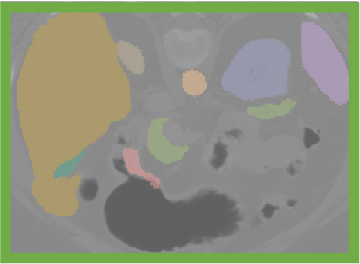}
    \footnotesize \textbf{Swin-UNETR}
  \end{minipage}
  \begin{minipage}{0.13\textwidth}
      \centering
   \includegraphics[height=2.2cm, width=\linewidth]{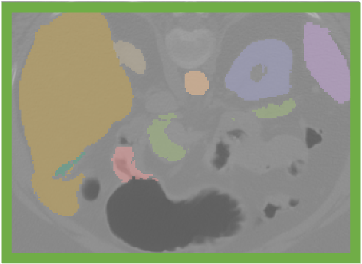}
   \footnotesize \textbf{UNETR}
  \end{minipage}
\begin{minipage}{0.13\textwidth}
      \centering
  \includegraphics[height=2.2cm, width=\linewidth]{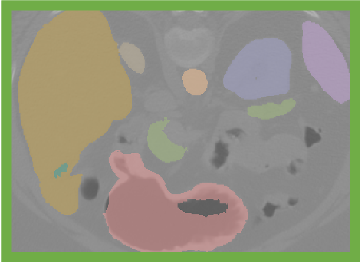}
  \footnotesize \textbf{Y-CH-Net}
  \end{minipage}
  \begin{minipage}{0.13\textwidth}
      \centering
    \includegraphics[height=2.2cm, width=\linewidth]{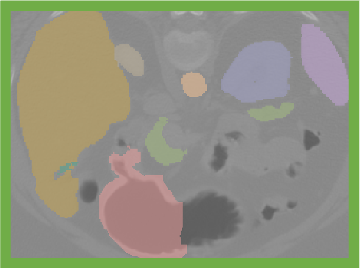}
    \footnotesize \textbf{Y-CT-Net(Swin)}
  \end{minipage}
  \begin{minipage}{0.13\textwidth}
      \centering
   \includegraphics[height=2.2cm, width=\linewidth]{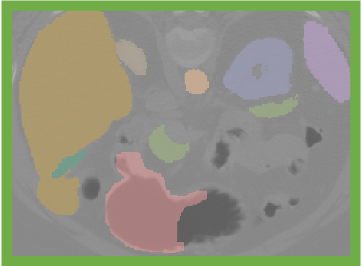}
   \footnotesize \textbf{Y-CT-Net(ViT)}
  \end{minipage}
  
\caption{Qualitative results on multi-organ segmentation. We mainly compare the visual results of our Y-CA-NET models with three recent state-of-art attention based methods (nnFormer, Swin-UNETR, and UNETR). Visual results demonstrate that our model has not only performed accurate prediction but also preserve the organ information which is missed by all three transformer-based methods.}\label{fig:visualization-multi-organ}
\vspace{-0.5em}
\end{minipage}
\end{figure*}

{\small
\bibliographystyle{ieee_fullname}
\bibliography{egbib}

\begin{thebibliography}{10}\itemsep=-1pt

\bibitem{antonelli2022medical}
Michela Antonelli, Annika Reinke, Spyridon Bakas, Keyvan Farahani, Annette Kopp-Schneider, Bennett~A Landman, Geert Litjens, Bjoern Menze, Olaf Ronneberger, Ronald~M Summers, et~al.
\newblock The medical segmentation decathlon.
\newblock {\em Nature communications}, 13(1):1--13, 2022.

\bibitem{Baid2021TheRB}
Ujjwal Baid, Satyam Ghodasara, Michel Bilello, Suyash Mohan, Evan Calabrese, Errol Colak, Keyvan Farahani, and Jayashree Kalpathy-Cramer.
\newblock The rsna-asnr-miccai brats 2021 benchmark on brain tumor segmentation and radiogenomic classification.
\newblock {\em ArXiv}, abs/2107.02314, 2021.

\bibitem{bardes2022vicregl}
Adrien Bardes, Jean Ponce, and Yann LeCun.
\newblock Vicregl: Self-supervised learning of local visual features.
\newblock {\em arXiv preprint arXiv:2210.01571}, 2022.

\bibitem{cao2021swin}
Hu Cao, Yueyue Wang, Joy Chen, Dongsheng Jiang, Xiaopeng Zhang, Qi Tian, and Manning Wang.
\newblock Swin-unet: Unet-like pure transformer for medical image segmentation.
\newblock {\em arXiv preprint arXiv:2105.05537}, 2021.

\bibitem{cao2021swinunet}
Hu Cao, Yueyue Wang, Joy Chen, Dongsheng Jiang, Xiaopeng Zhang, Qi Tian, and Manning Wang.
\newblock Swin-unet: Unet-like pure transformer for medical image segmentation.
\newblock {\em arXiv preprint arXiv:2105.05537}, 2021.

\bibitem{chang2021transclaw}
Yao Chang, Hu Menghan, Zhai Guangtao, and Zhang Xiao-Ping.
\newblock Transclaw u-net: Claw u-net with transformers for medical image segmentation.
\newblock {\em arXiv preprint arXiv:2107.05188}, 2021.

\bibitem{chen2021transunet}
Jieneng Chen, Yongyi Lu, Qihang Yu, Xiangde Luo, Ehsan Adeli, Yan Wang, Le Lu, Alan~L Yuille, and Yuyin Zhou.
\newblock Transunet: Transformers make strong encoders for medical image segmentation.
\newblock {\em arXiv preprint arXiv:2102.04306}, 2021.

\bibitem{dai2021transmed}
Yin Dai, Yifan Gao, and Fayu Liu.
\newblock Transmed: Transformers advance multi-modal medical image classification.
\newblock {\em Diagnostics}, 11(8):1384, 2021.

\bibitem{resunet}
Foivos~I Diakogiannis, Fran{\c{c}}ois Waldner, Peter Caccetta, and Chen Wu.
\newblock Resunet-a: A deep learning framework for semantic segmentation of remotely sensed data.
\newblock {\em ISPRS Journal of Photogrammetry and Remote Sensing}, 162:94--114, 2020.

\bibitem{dosovitskiy2020image}
Alexey Dosovitskiy, Lucas Beyer, Alexander Kolesnikov, Dirk Weissenborn, Xiaohua Zhai, Thomas Unterthiner, Mostafa Dehghani, Matthias Minderer, Georg Heigold, Sylvain Gelly, et~al.
\newblock An image is worth 16x16 words: Transformers for image recognition at scale.
\newblock {\em arXiv preprint arXiv:2010.11929}, 2020.

\bibitem{gani2022train}
Hanan Gani, Muzammal Naseer, and Mohammad Yaqub.
\newblock How to train vision transformer on small-scale datasets?
\newblock {\em arXiv preprint arXiv:2210.07240}, 2022.

\bibitem{han2021demystifying}
Qi Han, Zejia Fan, Qi Dai, Lei Sun, Ming-Ming Cheng, Jiaying Liu, and Jingdong Wang.
\newblock Demystifying local vision transformer: Sparse connectivity, weight sharing, and dynamic weight.
\newblock {\em arXiv preprint arXiv:2106.04263}, 2021.

\bibitem{hatamizadeh2022swin}
Ali Hatamizadeh, Vishwesh Nath, Yucheng Tang, Dong Yang, Holger~R Roth, and Daguang Xu.
\newblock Swin unetr: Swin transformers for semantic segmentation of brain tumors in mri images.
\newblock In {\em International MICCAI Brainlesion Workshop}, pages 272--284. Springer, 2022.

\bibitem{hatamizadeh2022unetr}
Ali Hatamizadeh, Yucheng Tang, Vishwesh Nath, Dong Yang, Andriy Myronenko, Bennett Landman, Holger~R Roth, and Daguang Xu.
\newblock Unetr: Transformers for 3d medical image segmentation.
\newblock In {\em Proceedings of the IEEE/CVF Winter Conference on Applications of Computer Vision}, pages 574--584, 2022.

\bibitem{resnet}
Kaiming He, Xiangyu Zhang, Shaoqing Ren, and Jian Sun.
\newblock Deep residual learning for image recognition.
\newblock In {\em Proceedings of the IEEE conference on computer vision and pattern recognition}, pages 770--778, 2016.

\bibitem{Unet3+}
Huimin Huang, Lanfen Lin, Ruofeng Tong, Hongjie Hu, Qiaowei Zhang, Yutaro Iwamoto, Xianhua Han, Yen-Wei Chen, and Jian Wu.
\newblock Unet 3+: A full-scale connected unet for medical image segmentation.
\newblock In {\em ICASSP 2020-2020 IEEE International Conference on Acoustics, Speech and Signal Processing (ICASSP)}, pages 1055--1059, 2020.

\bibitem{huang2021missformer}
Xiaohong Huang, Zhifang Deng, Dandan Li, and Xueguang Yuan.
\newblock Missformer: An effective medical image segmentation transformer.
\newblock {\em arXiv preprint arXiv:2109.07162}, 2021.

\bibitem{nnunet}
Fabian Isensee, Paul~F Jaeger, Simon~AA Kohl, Jens Petersen, and Klaus~H Maier-Hein.
\newblock nnu-net: a self-configuring method for deep learning-based biomedical image segmentation.
\newblock {\em Nature methods}, 18(2):203--211, 2021.

\bibitem{khan2022transformers}
Salman Khan, Muzammal Naseer, Munawar Hayat, Syed~Waqas Zamir, Fahad~Shahbaz Khan, and Mubarak Shah.
\newblock Transformers in vision: A survey.
\newblock {\em ACM computing surveys (CSUR)}, 54(10s):1--41, 2022.

\bibitem{landman2015miccai}
Bennett Landman, Zhoubing Xu, J Igelsias, Martin Styner, T Langerak, and Arno Klein.
\newblock Miccai multi-atlas labeling beyond the cranial vault--workshop and challenge.
\newblock In {\em Proc. MICCAI Multi-Atlas Labeling Beyond Cranial Vault—Workshop Challenge}, volume~5, page~12, 2015.

\bibitem{swin2021}
Ze Liu, Yutong Lin, Yue Cao, Han Hu, Yixuan Wei, Zheng Zhang, Stephen Lin, and Baining Guo.
\newblock Swin transformer: Hierarchical vision transformer using shifted windows.
\newblock In {\em Proceedings of the IEEE/CVF International Conference on Computer Vision}, pages 10012--10022, 2021.

\bibitem{liu2021swin}
Ze Liu, Yutong Lin, Yue Cao, Han Hu, Yixuan Wei, Zheng Zhang, Stephen Lin, and Baining Guo.
\newblock Swin transformer: Hierarchical vision transformer using shifted windows.
\newblock In {\em Proceedings of the IEEE/CVF International Conference on Computer Vision}, pages 10012--10022, 2021.

\bibitem{convnext}
Zhuang Liu, Hanzi Mao, Chao-Yuan Wu, Christoph Feichtenhofer, Trevor Darrell, and Saining Xie.
\newblock A convnet for the 2020s.
\newblock In {\em Proceedings of the IEEE/CVF Conference on Computer Vision and Pattern Recognition}, pages 11976--11986, 2022.

\bibitem{FCN}
Jonathan Long, Evan Shelhamer, and Trevor Darrell.
\newblock Fully convolutional networks for semantic segmentation.
\newblock In {\em Proceedings of the IEEE conference on computer vision and pattern recognition}, pages 3431--3440, 2015.

\bibitem{loshchilov2017decoupled}
Ilya Loshchilov and Frank Hutter.
\newblock Decoupled weight decay regularization.
\newblock {\em arXiv preprint arXiv:1711.05101}, 2017.

\bibitem{segresnet}
Andriy Myronenko.
\newblock 3d mri brain tumor segmentation using autoencoder regularization.
\newblock In {\em International MICCAI Brainlesion Workshop}, pages 311--320. Springer, 2018.

\bibitem{naseer2021intriguing}
Muhammad~Muzammal Naseer, Kanchana Ranasinghe, Salman~H Khan, Munawar Hayat, Fahad Shahbaz~Khan, and Ming-Hsuan Yang.
\newblock Intriguing properties of vision transformers.
\newblock {\em Advances in Neural Information Processing Systems}, 34:23296--23308, 2021.

\bibitem{rattunet}
Zhen-Liang Ni, Gui-Bin Bian, Xiao-Hu Zhou, Zeng-Guang Hou, Xiao-Liang Xie, Chen Wang, Yan-Jie Zhou, Rui-Qi Li, and Zhen Li.
\newblock Raunet: Residual attention u-net for semantic segmentation of cataract surgical instruments.
\newblock In {\em International Conference on Neural Information Processing}, pages 139--149. Springer, 2019.

\bibitem{paszke2019pytorch}
Adam Paszke, Sam Gross, Francisco Massa, Adam Lerer, James Bradbury, Gregory Chanan, Trevor Killeen, Zeming Lin, Natalia Gimelshein, Luca Antiga, et~al.
\newblock Pytorch: An imperative style, high-performance deep learning library.
\newblock {\em Advances in neural information processing systems}, 32, 2019.

\bibitem{ranasinghe2021selfsupervised}
Kanchana Ranasinghe, Muzammal Naseer, Salman Khan, Fahad~Shahbaz Khan, and Michael Ryoo.
\newblock Self-supervised video transformer, 2021.

\bibitem{unet}
Olaf Ronneberger, Philipp Fischer, and Thomas Brox.
\newblock U-net: Convolutional networks for biomedical image segmentation.
\newblock In {\em International Conference on Medical image computing and computer-assisted intervention}, pages 234--241, 2015.

\bibitem{saeed2022tmss}
Numan Saeed, Ikboljon Sobirov, Roba Al~Majzoub, and Mohammad Yaqub.
\newblock Tmss: An end-to-end transformer-based multimodal network for segmentation and survival prediction.
\newblock In {\em International Conference on Medical Image Computing and Computer-Assisted Intervention}, pages 319--329. Springer, 2022.

\bibitem{attunet}
Jo Schlemper, Ozan Oktay, Michiel Schaap, Mattias Heinrich, Bernhard Kainz, Ben Glocker, and Daniel Rueckert.
\newblock Attention gated networks: Learning to leverage salient regions in medical images.
\newblock {\em Medical image analysis}, 53:197--207, 2019.

\bibitem{shamshad2022transformers}
Fahad Shamshad, Salman Khan, Syed~Waqas Zamir, Muhammad~Haris Khan, Munawar Hayat, Fahad~Shahbaz Khan, and Huazhu Fu.
\newblock Transformers in medical imaging: A survey.
\newblock {\em arXiv preprint arXiv:2201.09873}, 2022.

\bibitem{Sharif_2022_BMVC}
Muhammad~Hamza Sharif, Dmitry Demidov, Asif Hanif, Mohammad Yaqub, and Min Xu.
\newblock Transresnet: Integrating the strengths of vits and cnns for high resolution medical image segmentation via feature grafting.
\newblock In {\em 33rd British Machine Vision Conference 2022, {BMVC} 2022, London, UK, November 21-24, 2022}. {BMVA} Press, 2022.

\bibitem{tang2022self}
Yucheng Tang, Dong Yang, Wenqi Li, Holger~R Roth, Bennett Landman, Daguang Xu, Vishwesh Nath, and Ali Hatamizadeh.
\newblock Self-supervised pre-training of swin transformers for 3d medical image analysis.
\newblock In {\em Proceedings of the IEEE/CVF Conference on Computer Vision and Pattern Recognition}, pages 20730--20740, 2022.

\bibitem{convmixer}
Asher Trockman and J~Zico Kolter.
\newblock Patches are all you need?
\newblock {\em arXiv preprint arXiv:2201.09792}, 2022.

\bibitem{valanarasu2021medical}
Jeya Maria~Jose Valanarasu, Poojan Oza, Ilker Hacihaliloglu, and Vishal~M Patel.
\newblock Medical transformer: Gated axial-attention for medical image segmentation.
\newblock In {\em Medical Image Computing and Computer Assisted Intervention--MICCAI 2021: 24th International Conference, Strasbourg, France, September 27--October 1, 2021, Proceedings, Part I 24}, pages 36--46. Springer, 2021.

\bibitem{vaswani2017attention}
Ashish Vaswani, Noam Shazeer, Niki Parmar, Jakob Uszkoreit, Llion Jones, Aidan~N Gomez, {\L}ukasz Kaiser, and Illia Polosukhin.
\newblock Attention is all you need.
\newblock {\em Advances in neural information processing systems}, 30, 2017.

\bibitem{mtunet}
Hongyi Wang, Shiao Xie, Lanfen Lin, Yutaro Iwamoto, Xian-Hua Han, Yen-Wei Chen, and Ruofeng Tong.
\newblock Mixed transformer u-net for medical image segmentation.
\newblock In {\em ICASSP 2022-2022 IEEE International Conference on Acoustics, Speech and Signal Processing (ICASSP)}, pages 2390--2394. IEEE, 2022.

\bibitem{nonlocal}
Xiaolong Wang, Ross Girshick, Abhinav Gupta, and Kaiming He.
\newblock Non-local neural networks.
\newblock In {\em Proceedings of the IEEE conference on computer vision and pattern recognition}, pages 7794--7803, 2018.

\bibitem{resnext}
Saining Xie, Ross Girshick, Piotr Doll{\'a}r, Zhuowen Tu, and Kaiming He.
\newblock Aggregated residual transformations for deep neural networks.
\newblock In {\em Proceedings of the IEEE conference on computer vision and pattern recognition}, pages 1492--1500, 2017.

\bibitem{xu2021levit}
Guoping Xu, Xingrong Wu, Xuan Zhang, and Xinwei He.
\newblock Levit-unet: Make faster encoders with transformer for medical image segmentation.
\newblock {\em arXiv preprint arXiv:2107.08623}, 2021.

\bibitem{zamir2022restormer}
Syed~Waqas Zamir, Aditya Arora, Salman Khan, Munawar Hayat, Fahad~Shahbaz Khan, and Ming-Hsuan Yang.
\newblock Restormer: Efficient transformer for high-resolution image restoration.
\newblock In {\em Proceedings of the IEEE/CVF Conference on Computer Vision and Pattern Recognition}, pages 5728--5739, 2022.

\bibitem{zhang2021transfuse}
Yundong Zhang, Huiye Liu, and Qiang Hu.
\newblock Transfuse: Fusing transformers and cnns for medical image segmentation.
\newblock In {\em Medical Image Computing and Computer Assisted Intervention--MICCAI 2021: 24th International Conference, Strasbourg, France, September 27--October 1, 2021, Proceedings, Part I 24}, pages 14--24. Springer, 2021.

\bibitem{zhou2018learning}
Chenhong Zhou, Shengcong Chen, Changxing Ding, and Dacheng Tao.
\newblock Learning contextual and attentive information for brain tumor segmentation.
\newblock In {\em International MICCAI brainlesion workshop}, pages 497--507. Springer, 2018.

\bibitem{zhou2021nnformer}
Hong-Yu Zhou, Jiansen Guo, Yinghao Zhang, Lequan Yu, Liansheng Wang, and Yizhou Yu.
\newblock nnformer: Interleaved transformer for volumetric segmentation.
\newblock {\em arXiv preprint arXiv:2109.03201}, 2021.

\bibitem{unet++}
Zongwei Zhou, Md~Mahfuzur Rahman~Siddiquee, Nima Tajbakhsh, and Jianming Liang.
\newblock Unet++: A nested u-net architecture for medical image segmentation.
\newblock In {\em Deep learning in medical image analysis and multimodal learning for clinical decision support}, pages 3--11. 2018.

\end{thebibliography}
}

\end{document}